\documentclass[11pt]{article}

\usepackage[final]{acl}

\usepackage{times}
\usepackage{latexsym}

\usepackage[T1]{fontenc}

\usepackage[utf8]{inputenc}

\usepackage{microtype} 

\usepackage{inconsolata}

\usepackage{graphicx}

\usepackage{algorithm}
\usepackage{algorithmic}
\usepackage{longtable}
\usepackage{tcolorbox}
\usepackage{stackengine}
\usepackage[T1]{fontenc}
\usepackage[accsupp]{axessibility}
\usepackage{caption}
\usepackage{enumitem}
\usepackage{marvosym}
\usepackage{booktabs}
\usepackage{pifont} 
\usepackage{float} 
\usepackage{placeins}
\usepackage[table]{xcolor}        
\usepackage{makecell}

\usepackage{scalerel} 
\usepackage{hyperref}

\usepackage{mathtools}
\usepackage{amsthm}
\usepackage{amsmath}

\usepackage{csquotes}

\usepackage{array}

\usepackage{tikz}
\usepackage[edges]{forest}

\definecolor{bgColor}{RGB}{252, 248, 242}     
\definecolor{dataColor}{RGB}{255, 242, 230}
\definecolor{timeColor}{RGB}{232, 243, 232}
\definecolor{archiColor}{RGB}{240, 232, 243}
\definecolor{decodeColor}{RGB}{235, 242, 252}
\definecolor{freeColor}{RGB}{222, 240, 248}

\usepackage{tabularx}

\usepackage{multirow}

\theoremstyle{definition}

\usepackage{todonotes}    

\title{Multimodal Unlearning Across Vision, Language, Video, and Audio: Survey of Methods, Datasets, and Benchmarks}

\author{
    Nobin Sarwar\hspace{.1em}\scalerel*{\includegraphics{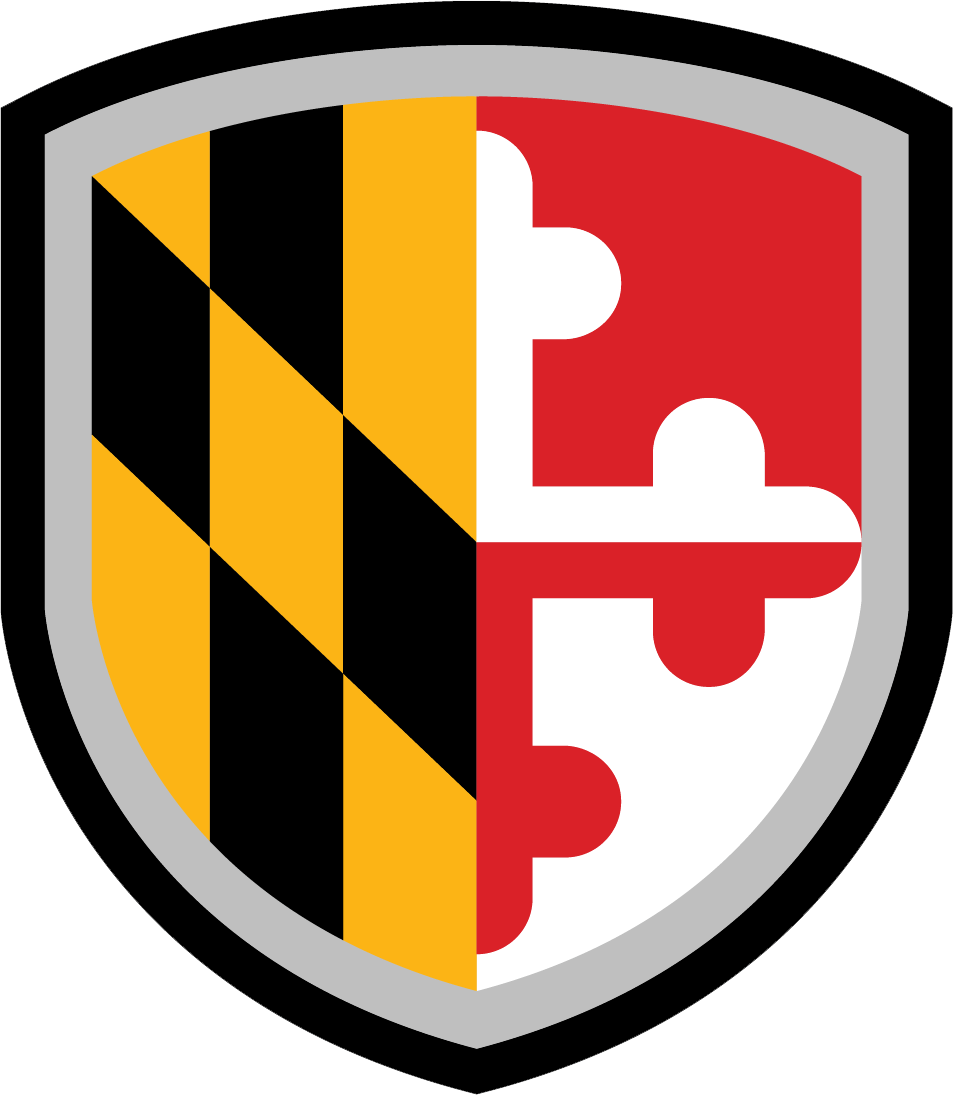}}{\textbf{O}},
    Shubhashis Roy Dipta\hspace{.1em}\scalerel*{\includegraphics{images/UMBC_logo.png}}{\textbf{O}},
    Zheyuan Liu\hspace{.1em}\scalerel*{\includegraphics{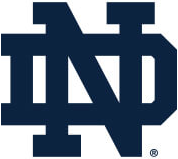}}{\textbf{O}},
    Vaidehi Patil\hspace{.1em}\scalerel*{\includegraphics{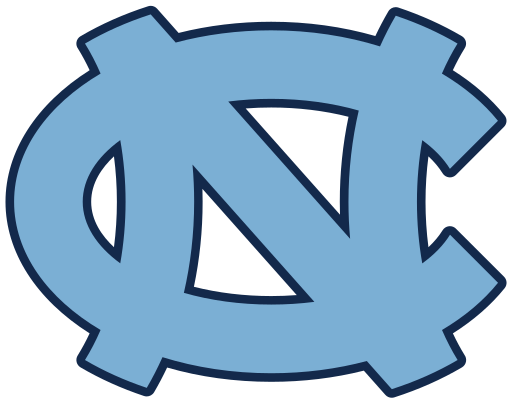}}{\textbf{O}} \\
    \hspace{.2em}\scalerel*{\includegraphics{images/UMBC_logo.png}}{\textbf{O}}
    \textbf{University of Maryland, Baltimore County}\\
    \hspace{.4em}\scalerel*{\includegraphics{images/UD_logo.png}}{\textbf{O}}
    \textbf{University of Notre Dame}
    \hspace{.4em}\scalerel*{\includegraphics{images/UNC_logo.png}}{\textbf{O}}
    \textbf{UNC Chapel Hill} \\
    \texttt{ \{sms2, sroydip1\}@umbc.edu}\\ 
    \texttt{zliu29@nd.edu, vaidehi@cs.unc.edu}
}

\begin{document}
\maketitle
\begin{abstract}
With the growing adoption of VLMs, DMs, LLMs, and AFMs, these multimodal foundation models can inadvertently encode sensitive, copyrighted, biased, or unsafe cross-modal associations that originate from their training data. Retraining after deletion requests or policy updates is often impractical, and targeted forgetting remains difficult because knowledge is distributed across shared representations. Multimodal unlearning addresses this challenge by enabling selective removal across modalities while retaining overall utility. This survey offers a unified, system-oriented view of multimodal unlearning across vision, language, audio, and video, grounded in recent advances, emerging applications, and open problems. Our taxonomy enables systematic comparison across model architectures and modalities, clarifying trade-offs among deletion strength, retention, efficiency, reversibility, and robustness. This survey highlights open problems and practical considerations to support future research and deployment of multimodal unlearning. We release a curated repository.\footnote{\url{https://smsnobin77.github.io/Awesome-Multimodal-Unlearning/}}
\end{abstract}

\section{Introduction}

Multimodal foundation models, including Vision Language Models (VLMs), Diffusion Models (DMs), Large Language Models (LLMs) and Audio Foundation Models (AFMs)-based~\cite{ho2020denoising, team2023gemini, yang2025qwen3, chu2023qwen, huang2024audiogpt} generators, support image, text, video, and audio understanding and generation at scale. Training on web-scale multimodal data improves generalization, but it can also induce memorization and undesired associations involving sensitive, copyrighted, biased, or unsafe content across modalities. As a result, deployed models may need to forget specific items or concepts, such as a copyrighted artwork, a private face, or a harmful trope, while retaining performance on the remaining data~\cite{fan2023salun, gandikota2023erasing, zhang2024unlearncanvas, sun2024unseg, chen2025safety, chen2025dual, facchiano2025video}. When deletion requests or policy updates affect only part of the training signal, retraining from scratch is often impractical~\cite{voigt2017eu, goldman2020introduction}. Targeted removal is challenging because knowledge is distributed in shared representations, so eliminating one association can disrupt unrelated behavior.

\begin{figure}[!t]
    \centering
    \includegraphics[width=1.0\columnwidth]{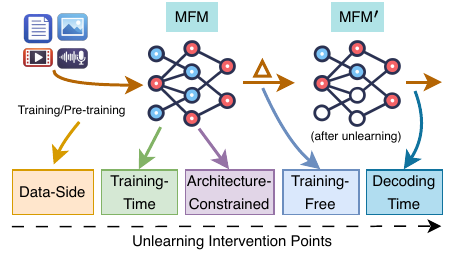}
    \caption{Unlearning intervention points for a Multimodal Foundation Model (MFM). Methods intervene at the data side, during training, via architecture-constrained edits, or at decoding time, producing an updated model (MFM$'$) with reduced influence from targeted content. Training-free methods use closed-form parameter or representation edits (denoted by $\Delta$) to directly transform the model without retraining.}
    \label{fig:mfm_treasure}
\end{figure}

\begin{table}[!t]
    \centering
    \resizebox{\columnwidth}{!}{
        \begin{tabular}{@{}lcccccc@{}}
        \toprule
        \textbf{Survey} & \textbf{Venue \& Year} & \textbf{System-first} & \textbf{Text} & \textbf{Image} & \textbf{Video} & \textbf{Audio} \\
        \midrule
        \citealp{si2023knowledge}                & arXiv'23 &        & \ding{52} &        &        &        \\
        \citealp{liu2024machine}              & arXiv'24 & \ding{52} & \ding{52} & \ding{52} &     &        \\
        \citealp{blanco2025digital}      & AIR'25   & \ding{52} & \ding{52} &        &        &        \\
        \citealp{liu2025rethinking}                   & NMI'25   &        & \ding{52} &      &        &        \\
        \citealp{feng2025survey}    & arXiv'25 & \ding{52} & \ding{52} & \ding{52} &     & \ding{52} \\
        \citealp{geng2025comprehensive}     & arXiv'25 & \ding{52} & \ding{52} & \ding{52} &     &        \\
        \midrule
        \textbf{Ours}           & \textbf{ACL'26} & \ding{52} & \ding{52} & \ding{52} & \ding{52} & \ding{52} \\
        \bottomrule
        \end{tabular}
    }
    \caption{Comparison of multimodal unlearning surveys across \textbf{modalities} and \textbf{system-first taxonomy} coverage.}
    \vspace{-10pt}
    \label{tab:survey_comparison_modalities}
\end{table}

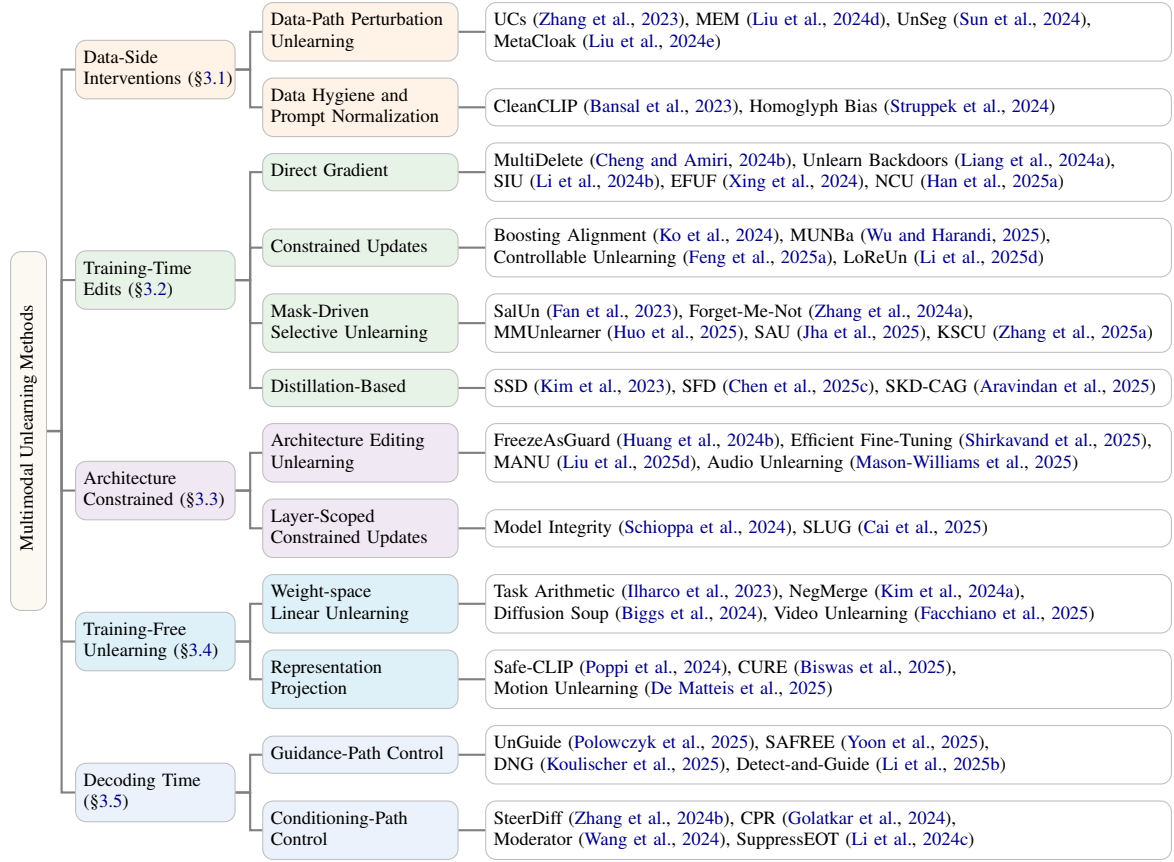
\begin{figure*}[t]
    \centering
    {\footnotesize
        \begin{forest}
            forked edges,
            for tree={
                grow=east,
                reversed=true,
                anchor=base west,
                parent anchor=east,
                child anchor=west,
                base=left,
                font=\scriptsize,
                rectangle,
                draw=gray!50,
                rounded corners,
                align=left,
                minimum width=5em,
                edge+={black!50, line width=0.8pt},
                s sep=6pt,
                inner xsep=3pt,
                inner ysep=3pt,
            },
            where level=0{minimum width=1.5em, minimum height=15em, align=center, font=\scriptsize, inner xsep=2pt}{},
            where level=1{text width=6em, font=\scriptsize}{},
            where level=2{text width=7.5em, font=\scriptsize}{},
            where level=3{text width=28em, font=\scriptsize}{},
            [
                {\rotatebox{90}{Multimodal Unlearning Methods}}, fill=bgColor 
                [
                    {Data-Side\\Interventions (\S\ref{subsec:data‑side-interventions})}, fill=dataColor
                    [ {Data-Path Perturbation\\Unlearning}, fill=dataColor 
                        [ {UCs~\cite{zhang2023unlearnable}, MEM~\cite{liu2024multimodal}, UnSeg~\cite{sun2024unseg},\\ MetaCloak~\cite{liu2024metacloak}} ]
                    ]
                    [ {Data Hygiene and \\Prompt Normalization}, fill=dataColor 
                        [ {CleanCLIP~\cite{bansal2023cleanclip}, Homoglyph Bias~\cite{struppek2024exploiting}} ]
                    ]
                ]
                [
                    {Training-Time\\Edits (\S\ref{subsec:training‑time-edits})}, fill=timeColor
                    [ {Direct Gradient}, fill=timeColor 
                        [ {MultiDelete~\cite{cheng2024multidelete}, Unlearn Backdoors~\cite{liang2024unlearning},\\ SIU~\cite{li2024single}, EFUF~\cite{xing2024efuf}, NCU~\cite{han2025unlearning}} ]
                    ]
                    [ {Constrained Updates}, fill=timeColor 
                        [ {Boosting Alignment~\cite{ko2024boosting}, MUNBa~\cite{wu2025munba},\\ Controllable Unlearning~\cite{fengcontrollable2025}, LoReUn~\cite{li2025loreun}} ]
                    ]
                    [ {Mask-Driven\\Selective Unlearning}, fill=timeColor 
                        [ {SalUn~\cite{fan2023salun},  Forget-Me-Not~\cite{zhang2024forget},\\ MMUnlearner~\cite{huo2025mmunlearner}, SAU~\cite{jha2025backdoor}, KSCU~\cite{zhang2025concept}} ]
                    ]
                    [ {Distillation-Based}, fill=timeColor 
                        [ {SSD~\cite{kim2023towards}, SFD~\cite{chen2025score}, SKD-CAG~\cite{aravindan2025sealing}} ]
                    ]
                ]
                [
                    {Architecture\\ Constrained (\S\ref{subsec:architecture-constrained})}, fill=archiColor
                    [ {Architecture Editing \\Unlearning}, fill=archiColor 
                        [ {FreezeAsGuard~\cite{huang2024freezeasguard}, Efficient Fine-Tuning~\cite{shirkavand2025efficient},  \\ MANU~\cite{liu2025modality}, Audio Unlearning~\cite{mason-williams2025machine}} ]
                    ]
                    [ {Layer-Scoped \\Constrained Updates}, fill=archiColor 
                        [ {Model Integrity~\cite{schioppa2024model}, SLUG~\cite{caitargeted2025}} ]
                    ]
                ]
                [
                    {Training-Free\\Unlearning (\S\ref{subsec:training‑free-unlearning})}, fill=freeColor
                    [ {Weight-space \\Linear Unlearning}, fill=freeColor 
                        [ {Task Arithmetic~\cite{ilharcoediting2023}, NegMerge~\cite{kim2024negmerge},\\ Diffusion Soup~\cite{biggs2024diffusion}, Video Unlearning~\cite{facchiano2025video}} ]
                    ]
                    [ {Representation\\ Projection}, fill=freeColor 
                        [ {Safe-CLIP~\cite{poppi2023removing}, CURE~\cite{biswas2025cure}, \\Motion Unlearning~\cite{de2025human}} ]
                    ]
                ]
                [
                    {Decoding Time\\ (\S\ref{subsec:decoding-time-unlearning})}, fill=decodeColor
                    [ {Guidance-Path Control}, fill=decodeColor 
                    [ {UnGuide~\cite{polowczyk2025unguide}, SAFREE~\cite{yoon2025safree}, \\DNG~\cite{koulischer2025dynamic}, Detect-and-Guide~\cite{li2025detect}} ]
                    ]
                    [ {Conditioning-Path \\Control}, fill=decodeColor 
                        [ {SteerDiff~\cite{zhang2024steerdiff}, CPR~\cite{golatkar2024cpr}, \\Moderator~\cite{wang2024moderator}, SuppressEOT~\cite{li2024get}} ]
                    ]
                ]  
            ]
        \end{forest}
    }
    \caption{Taxonomy of multimodal unlearning methods, organized by intervention stage and control pathway, with representative approaches in each category.}
    \vspace{-10pt}
    \label{fig:mmunlearn_methods_taxo}
\end{figure*}

These challenges have driven growing interest in multimodal unlearning as a mechanism for selective data removal and behavior correction. Early work on machine unlearning formalized the goal of removing training influence from learned models~\cite{cao2015towards, bourtoule2021machine}. Subsequent studies extend this objective to multimodal and generative systems, including DMs and VLMs, by enabling instance-level or concept-level deletion while preserving utility~\cite{kim2023towards, liu2024unlearning, li2024single, sun2024unseg, zhang2024forget, golatkar2024cpr}. These efforts make multimodal unlearning a central tool for model governance, supporting targeted forgetting without sacrificing utility.

While several surveys discuss multimodal unlearning (Table~\ref{tab:survey_comparison_modalities}), prior work often emphasizes unimodal settings such as text-only or image-only, or it restricts coverage to a narrow set of text-image systems. Many reviews also adopt algorithm-centric taxonomies organized around optimization objectives, which can obscure the intervention points that matter for deploying unlearning in end-to-end multimodal pipelines. As a result, the literature still lacks a unified exposition that connects mechanisms across vision, language, video, and audio.

Motivated by these gaps, this survey provides a comprehensive overview of multimodal unlearning for foundation models across vision, language, video, and audio. Instead of an algorithm-first taxonomy, we adopt a system-first view that organizes methods by intervention stage and control point, with \textbf{forgetting target scope} as the top-level split between \textbf{instance-level} and \textbf{concept-level} forgetting. This organization provides a stable scaffold for both established and emerging methods, enables cross-modal comparisons through shared control pathways, and clarifies trade-offs among deletion strength, utility retention, efficiency, and reversibility. This survey makes the following contributions to multimodal unlearning in foundation models:

\begin{itemize}[leftmargin=*, itemsep=-2pt, topsep=0pt]
  \item \textbf{Foundational Survey.} This survey synthesizes multimodal unlearning across foundation models for image, text, video, and audio, covering mechanisms, theory, and evaluation in one framework.

  \item \textbf{System-Level Lens.} We propose a system-first taxonomy organized by intervention stage and control pathway, enabling comparison across model classes and optimization families.
  
  \item \textbf{Emerging Frontiers.} We outline open challenges in evaluation, adversarial robustness, and deployment constraints, highlighting directions for accountable targeted unlearning.
\end{itemize}
\section{Formalizing Multimodal Unlearning}







The goal of multimodal unlearning is to remove the influence of a designated forget set while preserving utility on retained content across individual modalities and their shared representations~\cite{cao2015towards, ginart2019making, guo2020certified, bourtoule2021machine, li2024single}. Given a learning algorithm $A$ and multimodal training data $D = \{(I_i, T_i)\}_{i=1}^{N}$ consisting of paired images $I$ and texts $T$, let $M_o = A(D)$ denote the original model. For simplicity, we use image-text pairs; the formulation generalizes to video and audio. For a forget set $D_f \subseteq D$, define the retained data $D_r = D \setminus D_f$ and the retrained reference model $M_r = A(D_r)$. Single image unlearning corresponds to the setting $D_f = \{(I_f, T_f)\}$, where forgetting removes a single image-text association while preserving utility on $D_r$. Unlearning proceeds by applying $U$ to the original model and data to obtain $M_u = U(M_o, D, D_f)$. The unlearning objective requires the distribution induced by this procedure to be close to that of retraining, where closeness is measured over joint multimodal predictive outputs and model parameters through the induced distributions $P_r$ and $P_u$:
\[
P_r(A(D_r)) \approx P_u(U(M_o, D, D_f)).
\]
To formalize approximate retraining equivalence, an $(\varepsilon,\delta)$ unlearning criterion is adopted to provide theoretical guarantees and to mirror stability notions from Differential Privacy (DP)~\cite{dwork2006calibrating, sekhari2021remember, neel2021descent}:

{\footnotesize
\[
P\!\left[A(D \setminus D_f) \in R\right]
\le e^{\varepsilon}\, P\!\left[ U\!\left(A(D), D, D_f\right) \in R \right] + \delta,
\]
\[
P\!\left[ U\!\left(A(D), D, D_f\right) \in R \right]
\le e^{\varepsilon}\, P\!\left[ A(D \setminus D_f) \in R \right] + \delta,
\]
}

where $R$ ranges over measurable events in the joint space of model parameters and multimodal predictive outputs. The pair of inequalities defines a symmetric divergence bound, ensuring that retraining and unlearning induce distributions that are mutually close up to $(\varepsilon,\delta)$. Probabilities $P[\cdot]$ are taken over the randomness of $A$ and $U$ and any evaluation sampling, with $\varepsilon = \delta = 0$ recovering exact retraining equivalence.

\textbf{Optimization Objective.} In multimodal models, forgetting is operationalized through a two term objective that suppresses responses associated with the forget set while preserving utility on the retained set across individual modalities and their fusion mechanisms:
\[
\min_{\theta}\; J(\theta)
\;=\;
F_{\text{forget}}(\theta; D_f)
\;+\;
\lambda\,F_{\text{retain}}(\theta; D_r),
\]

where \(F_{\text{forget}}\) reduces the influence of multimodal associations in \(D_f\) and \(F_{\text{retain}}\) preserves utility on retained multimodal dataset \(D_r\). 

\subsection{Formulation of VLM Unlearning}
\label{app:vlm_formulation}






VLM unlearning targets the components that bind vision and language, supporting both instance-level and concept-level removal, while keeping unimodal competence intact. Let a VLM comprise a vision encoder $f_v$, a text encoder $f_t$, and a fusion head $F$. Given forget pairs $D_f=\{(x,c_f)\}$ that align an image $x$ with a forget concept prompt $c_f$ and retain pairs $D_r$, a compact objective balances suppression and utility:

\[
\begin{split}
\min_{\theta \in \{\theta_v,\theta_{\text{fusion}}\}}
\; L_{\text{retain}}(D_r;\theta)
\;+\;
\lambda\,L_{\text{forget}}(D_f;\theta) \\
\;+\;
\mu\,\Omega(\theta).
\end{split}
\]

A concept hinge decouples semantics by penalizing violations of \(S_{\theta}(x,c_f) \le m\) for \((x,c_f) \in D_f\), where \(m\) is a similarity threshold that sets the target upper bound on forget-pair similarity, while a consistency or caption term preserves performance on \(D_r\)~\cite{li2024single}. Selective updates use a saliency mask $S$ so that

\vspace{-10pt}

\[
\Delta \theta \;=\; -\,\eta\, S \odot \nabla_{\theta}\!\left( L_{\text{forget}} + \lambda\, L_{\text{retain}} \right),
\]




\subsection{Formulation of DM Unlearning}
\label{app:dms_formulation}

Diffusion Model unlearning focuses on the conditional denoising path tied to a target concept. Let $\epsilon_{\theta}(x_t,c,t)$ denote the denoiser with conditioning $c$. A teacher guided loss attenuates the target channel,

\vspace{-10pt}

\[
\begin{aligned}
L_{\text{forget}} &= \mathbb{E}\!\left[\left\| \epsilon_{\theta}(x_t,c_f,t) - \tilde{\epsilon}(x_t,t) \right\|_2^2\right],\\
L_{\text{retain}} &= \mathbb{E}\!\left[\left\| \epsilon(x_t,t) - \epsilon_{\theta}(x_t,c_r,t) \right\|_2^2\right],
\end{aligned}
\]

so $\epsilon_{\theta}$ aligns with an unconditional or safe teacher on $c_f$ while generation quality on $D_r$ remains stable~\cite{gandikota2023erasing, zhang2024forget}. Representation editing complements loss shaping by modifying cross-attention: keys and values associated with $c_f$ are mapped to neutral surrogates, implemented as low rank or sparse updates $W_{\text{attn}} \leftarrow W_{\text{attn}} - \alpha\, \Pi_{c_f}$ across timesteps~\cite{kumari2023ablating, gandikota2024unified}. Sampling time steering reduces classifier-free guidance $s$ or injects negative prompts to deflect $c_f$ without weight changes~\cite{zhang2024forget}. 

\section{Multimodal Unlearning Methods}
\label{sec:methods}
We organize multimodal unlearning methods by \textbf{forgetting target scope} and, within each scope, by the \textbf{intervention stage} and control mechanism in the multimodal pipeline (Figures~\ref{fig:mfm_treasure} and~\ref{fig:mmunlearn_methods_taxo}).

\subsection{Data‑Side Interventions}
\label{subsec:data‑side-interventions}

\textbf{Data-Path Perturbation Unlearning.} Data-path perturbation unlearning edits inputs, not weights, to reduce the learnability of targeted clusters, pairs, or subjects while preserving utility on the remaining corpus~\cite{zhang2023unlearnable, liu2024multimodal, sun2024unseg, liu2024metacloak}. Typical instantiations include cluster-wise perturbations, coupled image-text edits, segmentation-disrupting generators, and transformation-robust cloaks for personalization resistance. We view this as constrained perturbation design:
\[
\|p_{\mathrm{img}}(x)\| \le \epsilon_{\mathrm{img}},\quad
\|p_{\mathrm{txt}}(t)\| \le \epsilon_{\mathrm{txt}},\quad
p \in \Pi_{T},
\]
where $p$ perturbs target samples within image/text budgets and enforces robustness to common transforms $T$.

\textbf{Data Hygiene and Prompt Normalization.}
Data hygiene reduces backdoor and trigger effects by curating or down-weighting suspicious image-text pairs, while prompt normalization canonicalizes visually or lexically similar tokens prior to optimization~\cite{bansal2023cleanclip, struppek2024exploiting}. We summarize both operations as:


\[
w(x,t)\in[0,1], \qquad t \mapsto N(t),
\]
where $w(x,t)$ down-weights or removes flagged pairs and $N(\cdot)$ maps look-alike tokens or script variants to canonical forms. This abstraction highlights two complementary levers, corpus curation and prompt normalization, that mitigate spurious associations at the data and input levels.

\subsection{Training‑Time Edits}
\label{subsec:training‑time-edits}

 
\textbf{Direct Gradient.}
Direct gradient methods formulate unlearning as targeted risk minimization over a retain set and a forget set. The procedure first identifies behaviors to remove using curated data or token-level signals, then updates parameters so that responses on the forget set degrade while performance on retained data remains stable. In VLMs, this approach includes clean fine-tuning that disrupts poisoned cross-modal associations and objectives that decouple cross-modal structure from unimodal features~\cite{bansal2023cleanclip, cheng2024multidelete}. A generic objective used across contrastive and generative settings is:
\[
\begin{aligned}
J(\theta) =\;&
\mathbb{E}_{(x_r,y_r)\in R}\,\mathcal{L}_u(f_\theta(x_r),y_r) \nonumber \\
&+ \alpha\,\mathbb{E}_{(x_f,y_f)\in F}\,\mathcal{L}_f(f_\theta(x_f),y_f) \nonumber \\
&+ \beta\,\mathbb{E}_{x_f\in F}\,D_a(f_\theta,x_f;a) \nonumber \\
&+ \gamma\,\Omega(\theta,\theta_0),
\end{aligned}
\]
where the first term preserves utility on retained data, the second suppresses behavior on the forget set, the optional redirection term steers outputs away from forgotten content, and the regularizer limits deviation from a reference model.

Diffusion models instantiate this template through preference-aligned denoising, anchor redirection, or uncertainty-based objectives, while text-to-video variants apply similar updates to the shared text encoder~\cite{park2024direct, kumari2023ablating, li2024safegen, liu2024unlearning, spartalis2025unleashing}. Audio and music systems adapt the same principle with task-specific losses that reduce speaker identity evidence, suppress memorized transcripts, or remove licensed content while preserving generation quality~\cite{kim2025not, liu2025unlearning, pathak2025quantum, kim2025no}.


\textbf{Constrained Updates.}
Constrained update methods retain the locate-then-unlearn workflow but make the trade-off between forgetting and retention explicit. Instead of relying on unconstrained optimization, these approaches impose bounds that limit residual competence on the forget set and restrict deviation from a reference model while optimizing utility on retained data. At a high level, forgetting can be framed as constrained risk minimization~\cite{schioppa2024model, wu2025munba, fengcontrollable2025},
\[
\begin{aligned}
\min_{\theta}\quad 
& \underbrace{\mathcal{J}_R(\theta)}_{\text{retain risk}}
\;+\;
\underbrace{\Omega(\theta,\theta_0)}_{\text{stability}} \\[2pt]
\text{s.t.}\quad 
& \underbrace{\mathcal{C}_f(\theta)}_{\text{forget efficacy}} \le 0,\qquad
\underbrace{\mathcal{C}_i(\theta)}_{\text{integrity}} \le 0
\end{aligned}
\]
where the objective preserves performance on retained data through a stability prior, while the constraints enforce forgetting efficacy and model integrity relative to a reference checkpoint.

Existing methods differ primarily in how they instantiate these constraints and balance them during optimization. Joint constrained updates reconcile gradients for forgetting and utility~\cite{wu2025munba}. Integrity-aware formulations preserve perceptual similarity or enforce monotonic improvement across objectives~\cite{schioppa2024model, ko2024boosting}. Related work applies importance-weighted deletion, knowledge tracing that removes fine-grained classes while retaining coarse recognition, or constrained recommendation updates that track divergence under user-level deletions~\cite{alberti2025data, sinha2025multi, li2025loreun}. 

\textbf{Mask-Driven Selective Unlearning.} 
Mask-driven methods follow the locate-then-unlearn workflow but constrain updates to a localized support identified through saliency, attention, or architectural structure. By restricting modification to parameters, features, spatial regions, or selected diffusion steps that most strongly encode the forget signal, these methods focus optimization where it matters while limiting collateral effects on retained behavior. Representative approaches include parameter-level masks derived from gradient or Fisher saliency~\cite{fan2023salun, huo2025mmunlearner}, activation or spatial masks that suppress trigger-aligned attention~\cite{zhang2024forget, jha2025backdoor}, and diffusion-time masking schemes that update only a subset of denoising steps to stabilize multi-concept unlearning~\cite{zhang2025concept, li2025sculpting}.

\textbf{Distillation-Based Unlearning.}
Distillation-based unlearning follows the locate-then-unlearn paradigm by transferring behavior through a teacher-student setup, where the student is guided toward a safe target while retaining competence on non-forgotten prompts. Methods mainly differ in how the unlearning target is specified and how supervision is obtained. Existing work includes self-distillation that aligns conditional and unconditional predictions to suppress unsafe concepts~\cite{kim2023towards}, data-free distillation that relies on lightweight generators to approximate forget and retain distributions~\cite{chen2025score}, and attention-guided distillation that weakens adversarial trigger pathways during knowledge transfer~\cite{aravindan2025sealing}. Across settings, distillation provides a training-time mechanism to redirect model behavior without direct access to original training data, while controlling drift relative to a reference model.

\subsection{Architecture-Constrained Unlearning}
\label{subsec:architecture-constrained}

\textbf{Architecture Editing Unlearning.}
Architecture editing methods follow the locate-then-unlearn paradigm by modifying network structure through pruning, freezing, or controlled regrowth. Instead of reshaping the loss, these methods intervene directly in the computation graph to restrict pathways that encode the forget signal while limiting parameter drift elsewhere. Representative approaches include modality-aware pruning with light fine-tuning~\cite{liu2025modality}, bilevel pruning coupled with suppression objectives~\cite{shirkavand2025efficient}, freezing adaptation-critical tensors during downstream adaptation~\cite{huang2024freezeasguard}, and prune-and-regrow strategies in audio models that restore capacity before fine-tuning on retained data~\cite{mason-williams2025machine}. By confining updates to localized structural components, architecture editing can better preserve retained behavior than global parameter updates, although its success depends on precise localization of the forget signal and sufficient residual capacity in the remaining network.

\textbf{Layer-Scoped Constrained Updates.} Layer-scoped constrained updates follow locate-then-unlearn by first identifying where the target concept concentrates, then restricting edits to that support to limit collateral damage. SLUG~\cite{caitargeted2025} localizes the update to a selected layer to achieve targeted removal with minimal parameter drift. Model-integrity-controlled updates~\cite{schioppa2024model} instead constrain the update to preserve base behavior, typically by penalizing deviations from a reference model while enforcing forgetting efficacy.

\begin{table*}[t!]
    \centering\scriptsize
    \begin{tabular}{p{0.08\textwidth} p{0.30\textwidth} p{0.24\textwidth} p{0.26\textwidth}}
        \toprule
        \textbf{Modality} & \textbf{Dataset} & \textbf{Size} & \textbf{Used in} \\
        \midrule

        \rowcolor{gray!25} 
        \multicolumn{4}{c}{%
          \rule{0pt}{2.2ex}\textbf{Identity Unlearning}\rule[-.9ex]{0pt}{0pt}
        } \\
        \midrule

        \multirow{16}{*}{Image}
            & CelebA~\cite{liu2015deep}            & 202,599 images & \citealp{dai2023training, dontsov2024clear, huang2024enhancing, caitargeted2025, zhang2024learning, liu2025protecting} \\
            \cmidrule(lr){2-4}
            
            & CelebA-HQ~\cite{karras2018progressive}         & 30K high-quality images from CelebA & \citealp{huang2024enhancing, alberti2025data, nagasubramaniam2025prompting} \\
            \cmidrule(lr){2-4}
            
            & Flickr-Faces HQ~\cite{karras2019style}   & 70K face images & \citealp{nagasubramaniam2025prompting} \\
            \cmidrule(lr){2-4}

            & CASIA-WebFace~\cite{yi2014learning}   & 494K face images  & \citealp{dontsov2024clear} \\
            \cmidrule(lr){2-4}
            
            & FairFace~\cite{karkkainen2021fairface}         & 108,501 face images & \citealp{alabdulmohsinclip2024} \\
            \cmidrule(lr){2-4}
            
            
            
            & MillionCelebs~\cite{zhang2020global}     & 18.8M images of 636K identities & \citealp{dontsov2024clear} \\
            \cmidrule(lr){2-4}
            
            & VGGFace2~\cite{cao2018vggface2}          & 3.3M face images & \citealp{liu2024metacloak, li2025towards} \\
            \cmidrule(lr){2-4}
            
            & PinsFaces~\cite{pinsfacerecognition2020}         & 17.5K cropped face photos & \citealp{kravets2025zero, kravets2025clip} \\
            
            
            

        \midrule



        \multirow{1}{*}{Audio}
            & VoxCeleb1~\cite{nagrani2017voxceleb}       & 150K utterances from 1.3k speakers & \citealp{cheng2025speech} \\
        \midrule

        \rowcolor{gray!25} 
        \multicolumn{4}{c}{%
          \rule{0pt}{2.2ex}\textbf{Affect and Video Unlearning}\rule[-.9ex]{0pt}{0pt}
        } \\
        \midrule

        \multirow{3}{*}{Image}
             & EmoSet~\cite{yang2023emoset} & 3.3M images, 118K human-labeled with emotion and attributes. & \citealp{zhou2024visual} \\
             \cmidrule(lr){2-4}
             
             
             & UnBiasedEmo~\cite{panda2018contemplating} & 3K affective images (6 emotion classes) & \citealp{zhou2024visual} \\
        \midrule

        \multirow{1}{*}{Video}
             & UCF101~\cite{soomro2012ucf101} & 13K videos across 101 action classes & \citealp{cheng2024mu} \\


            
            
            
        \bottomrule

    \end{tabular}
    \caption{Key datasets commonly used in multimodal unlearning. Datasets are grouped by unlearning setting (identity unlearning; affect and video unlearning) and modality, with their sizes and representative studies. Additional dataset categories are provided in Tables~\ref{tab:person_copyright_datasets}, \ref{tab:speech_safety_datasets}, and \ref{tab:class_datasets} (App.~\ref{app:datasets}).}
    \vspace{-10pt}
    \label{tab:identity_affect_datasets}
\end{table*}

\subsection{Training‑Free Unlearning}
\label{subsec:training‑free-unlearning}

\textbf{Weight-Space Linear Unlearning.}
Weight-space Linear Unlearning (WLU) follows the locate-then-unlearn paradigm but replaces iterative optimization with closed-form edits in parameter space. Instead of retraining, these methods modify a reference checkpoint through linear operations that suppress unwanted behavior while largely preserving retained utility. Representative instances include task-vector subtraction or negation~\cite{ilharcoediting2023}, sign-consistent aggregation and weight negation~\cite{kim2024negmerge}, low-rank suppression updates derived from safe and unsafe activations~\cite{facchiano2025video}, and checkpoint averaging schemes that exclude shards associated with the forget data~\cite{biggs2024diffusion}.

Formally, WLU constructs an edited model $\theta'$ as a linear transformation of a reference model $\theta_{0}$, where the direction and magnitude of the update encode the target behavior to remove. These edits remain training-free, composable across tasks, and easy to reverse, which makes WLU attractive when retraining is infeasible or when rapid post hoc control is required.

\textbf{Representation Projection Unlearning.}
Representation Projection Unlearning (RPU) follows the locate-then-unlearn paradigm but replaces iterative optimization with closed-form edits in representation space. Instead of updating model parameters, these methods suppress target concepts by projecting internal activations or attention outputs away from a learned subspace associated with the forget signal. This strategy localizes change, limits collateral effects, and preserves overall model structure. Representative examples include CURE~\cite{biswas2025cure}, which projects joint embeddings to remove visual concepts, and related projection-based methods that operate on multimodal representation spaces~\cite{poppi2023removing, de2025human}. The core operation applies an orthogonal projection that removes components aligned with the forget subspace:
\[
h' \;=\; (I - U U^{\top})\,h,
\qquad
W' \;=\; W\,(I - U U^{\top}),
\]
where \(h\) denotes an intermediate representation, \(W\) an attention or projection matrix, and \(U\) a column-orthonormal basis spanning the forget subspace. The operator \(I - U U^{\top}\) filters out directions linked to the target concept, yielding edited representations or projections without retraining. The effectiveness of RPU depends on how accurately the forget subspace is identified. Existing methods estimate \(U\) by factorizing attention features or by analyzing joint embedding statistics, which enables targeted suppression while keeping unrelated representations intact.

\begin{table*}[t]
  \centering
  \small
  \resizebox{\textwidth}{!}{%
    \begin{tabular}{p{3.8cm} p{1.2cm} p{3cm} p{1.5cm} p{3.5cm} p{4cm}}
      \toprule
      \textbf{Benchmark} & \textbf{Modality} & \textbf{Unlearning Target} & \textbf{Task Type} & \textbf{Key Statistics} & \textbf{Evaluation Objective} \\
      \cmidrule(lr){1-6}

      \multicolumn{6}{c}{\textbf{Unified Benchmark Suites}} \\
      \cmidrule(lr){1-6}

      MU-Bench~\cite{cheng2024mu} & Multimodal & Mixed (instances, datasets, modalities) & Multi-task & 9 datasets, 20 architectures & Unified unlearning evaluation (efficacy, utility, efficiency) \\
      \addlinespace[0.35em]

      MLLMU-Bench~\cite{liu2025protecting} & VLM & Private data (fictitious \& real identities) & Multi-task QA & 500 fictitious and 153 public celebrities, 20.7K QA pairs & Privacy unlearning across efficacy, generalization, utility \\
      \addlinespace[0.35em]

      PEBench~\cite{xu2025pebench} & VLM & Synthetic identities \& events & Multi-task & 200 identities, 8K images, 16K QA pairs & Privacy and event unlearning with controlled scope and audits \\

      \addlinespace[0.35em]

      UMU-Bench~\cite{wang2025umu} & VLM & knowledge instances & Multi-task & 500 fictitious, 153 real & Modality-aligned unlearning completeness and utility \\

      \cmidrule(lr){1-6}
      \multicolumn{6}{c}{\textbf{Identity and Privacy Unlearning}} \\
      \cmidrule(lr){1-6}

      CLEAR~\cite{dontsov2024clear} & VLM & Identity & VQA & 200 synthetic IDs, 3.7K images, 4K QA pairs & Identity leakage reduction with VQA accuracy retention \\
      \addlinespace[0.35em]

      FIUBench~\cite{ma2024benchmarking} & VLM & Identity & VQA & 400 synthetic IDs, 8K QA pairs & Right-to-be-forgotten under privacy constraints \\
      \addlinespace[0.35em]

      UnSLU-BENCH~\cite{koudounas2025alexa} & Audio & Speaker & Intent classification & Multi-speaker data, 4 languages & Speaker erasure with intent accuracy retention \\

      \cmidrule(lr){1-6}
      \multicolumn{6}{c}{\textbf{Content and Knowledge Unlearning}} \\
      \cmidrule(lr){1-6}

      CPDM~\cite{ma2024dataset} & DM & Styles/portraits & Generation & 2.1K anchors, 18.9K generated images & Copyright similarity reduction with quality retention \\
      \addlinespace[0.35em]

      UnlearnCanvas~\cite{zhang2024unlearncanvas} & DM & Artistic styles & Generation & 60 styles, 20 objects, high-res stylized images & Style forgetting with retention and generation fidelity/diversity \\
      \addlinespace[0.35em]

      Holistic Unlearning~\cite{moon2024holistic} & DM  & Mixed concepts & Generation & 33 target concepts, 16k prompts per concept & Faithfulness, alignment, robustness, efficiency \\
      \addlinespace[0.35em]

      Six-CD~\cite{ren2025six} & DM & Concept removal & Generation & Six concept categories, dual-version prompts & Cross category concept suppression with retainability checks \\
      \addlinespace[0.35em]

      MMUBench~\cite{li2024single} & VLM  & Concept-level visual recognition & VQA & 20 concepts, 50 images per concept & Concept-level visual unlearning with multimodal utility retention \\
      \addlinespace[0.35em]

      UnLOK-VQA~\cite{patilunlearning2024} & VLM & Targeted pretrained multimodal knowledge & VQA & 500 samples with rephrase and neighborhood data & Privacy leakage reduction under attack-and-defense evaluation \\
      \addlinespace[0.35em]

      SafeEraser~\cite{chen2025safeeraser} & VLM & Harmful knowledge & VQA & 3K images, 28.8K QA pairs & Harmful response reduction while preserving VQA utility \\

      \bottomrule
    \end{tabular}%
  }
  \caption{Representative multimodal unlearning benchmarks grouped by unlearning target, reporting modality, task type, scale, and evaluation objective. Multimodal refers to image, text, audio, and video.}
  \vspace{-10pt}
  \label{tab:mm_unlearning_benchmarks}
\end{table*}

\subsection{Decoding Time Unlearning}
\label{subsec:decoding-time-unlearning}

\textbf{Guidance-Path Control.} Guidance-path control performs locate-then-unlearn at decoding time by modifying the sampler rather than the model parameters. Instead of updating weights, these methods reshape the score used during generation to suppress target concepts while preserving visual quality and stylistic coherence. The base checkpoint remains fixed, enabling prompt-time selectivity and compatibility with standard sampling procedures, as in Dynamic Negative Guidance~\cite{koulischer2025dynamic}, UnGuide~\cite{polowczyk2025unguide}, and Steering Guidance~\cite{park2025steering}, as well as detection-driven variants that combine concept identification with localized guidance to restrict unsafe content during generation~\cite{li2025detect, yoon2025safree}. A common formulation adjusts the predicted score at each denoising step:

\[
\hat{\epsilon}_t
= \epsilon_\theta(x_t,c)
+ a_t\bigl[\epsilon_{\mathrm{alt}}(x_t,c)-\epsilon_\theta(x_t,c)\bigr]
- b_t\, M_t\, d_t,
\]
where \(x_t\) denotes the latent at step \(t\), \(c\) the conditioning signal, and \(\epsilon_\theta\) the base predictor. The remaining terms introduce time-dependent steering, optional alternative guidance, and localized suppression through masks and direction vectors.

\textbf{Conditioning-Path Control.}
Conditioning-path control performs locate-then-unlearn by modifying the conditioning signal that guides generation, while leaving model parameters unchanged. The sampler therefore operates under a weakened or safer condition for the target concept, which preserves inference latency and supports reversible control~\cite{zhang2024steerdiff, li2024get, wang2024moderator, golatkar2024cpr, bui2025hiding}. 

Let \(c\) denote the original conditioning input, such as a text embedding or a retrieval-augmented vector, and let \(s_\theta\) be the conditional score used during sampling. Conditioning-path control constructs a transformed condition
\[
c' \;=\; (1 - \alpha)\,c \;+\; \alpha\, T(c, R, \text{policy}),
\]
and then applies \(s_\theta(x_t \mid c')\) at each denoising step. The scalar \(\alpha \in [0,1]\) controls the strength of intervention, \(R\) denotes an optional retrieval store, and \(T\) specifies the control mechanism.

Representative instantiations include projection toward a safe subspace in SteerDiff~\cite{zhang2024steerdiff}, policy-aware prompt rewriting and coordination in Moderator~\cite{wang2024moderator}, hidden-key conditioning that gates concept activation~\cite{bui2025hiding}, and retrieval mixing with selective deletion in CPR~\cite{golatkar2024cpr}. These approaches share a common structure that alters conditioning pathways to suppress targeted concepts without retraining.

\begin{figure}[t]
    \centering
    {\footnotesize
        \begin{forest}
            forked edges,
            for tree={
                grow=east,
                reversed=true,
                anchor=base west,
                parent anchor=east,
                child anchor=west,
                base=left,
                font=\scriptsize,
                rectangle,
                draw=gray!50,
                rounded corners,
                align=left,
                minimum width=4.5em,
                edge+={black!50, line width=0.7pt},
                s sep=4pt,
                inner xsep=2pt,
                inner ysep=2pt,
            },
            where level=0{minimum width=1.2em, minimum height=12.5em, align=center, font=\scriptsize, inner xsep=1.5pt}{},
            where level=1{text width=6.5em, font=\scriptsize}{},
            where level=2{text width=11.5em, font=\scriptsize}{},
            [
                {\rotatebox{90}{Unlearning Evaluation Frameworks}}
                [
                    {Forget Quality \\ and Safety (\S\ref{subsec:forget_quality})}, fill=archiColor
                    [ {UA, FA@K, Wasserstein distance, \\ CLIP-Cls Drop, CLIP-Sim Drop} ]
                ]
                [
                    {Safety and Content \\ Forgetting (\S\ref{subsec:safety_Content})}, fill=timeColor
                    [ {Refusal Rate, VLM-as-Judge, \\Inappropriate Content Rate} ]
                ]
                [
                    {Attack-Based \\ Privacy (\S\ref{subsec:attack_based_privacy})}, fill=decodeColor
                    [ {MIA, Identity Matching, \\SIM, spk-ZRF} ]
                ]
                [
                    {Utility and \\ Faithfulness (\S\ref{subsec:model_utility})}, fill=freeColor
                    [ {Top-k Acc, Recall@K, BLEU, \\CLIP Score, FID, LPIPS,\\  LLM-as-a-Judge, Human-as-Judge} ]
                ]
                [
                    {Adversarial \\ Robustness (\S\ref{subsec:adversarial_perturbation})}, fill=dataColor
                    [ {Attack Success Rate} ]
                ]
                [
                    {Compute and \\ Environment (\S\ref{subsec:compute_budget})}, fill=bgColor
                    [ {WCT, Memory, FLOPs, \\Energy, CO$_2$e} ]
                ]
            ]
        \end{forest}
    }
    \caption{Evaluation dimensions and representative metrics for multimodal unlearning. Details in App.~\ref{app:evaluation_frameworks}.}
    \label{fig:mmunlearn_eval_taxo}
    \vspace{-10pt}
\end{figure}
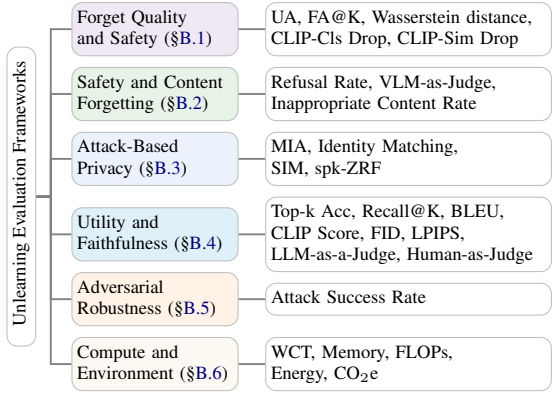

\vspace{-5pt}

\section{Datasets for Multimodal Unlearning}
\label{sec:Datasets_main}

We organize datasets for multimodal unlearning by application setting and modality, and summarize them across four tables. Table \ref{tab:identity_affect_datasets} covers identity, affect, and video unlearning benchmarks, including face, emotion, and action datasets. Table \ref{tab:person_copyright_datasets} focuses on personalization and copyright unlearning, capturing subject-specific and licensed content removal in generative models. Table \ref{tab:speech_safety_datasets} presents speech and safety robustness datasets used to study speaker, content, and jailbreak unlearning, along with web-scale data hygiene benchmarks that remove noisy or sensitive alignments from large pretraining corpora. Finally, Table \ref{tab:class_datasets} reports class-level unlearning benchmarks spanning image classification and segmentation settings (Tables~\ref{tab:person_copyright_datasets}, \ref{tab:speech_safety_datasets}, and \ref{tab:class_datasets} are in Appendix~\ref{app:datasets}).

\section{Multimodal Unlearning Benchmarks}
Multimodal unlearning has become central to addressing privacy, copyright, and safety concerns in vision-language and generative models. We review recent benchmarks that evaluate multimodal unlearning across diverse targets, modalities, and tasks. As summarized in Table~\ref{tab:mm_unlearning_benchmarks}, existing benchmarks range from unified suites spanning multiple datasets and architectures to task-specific evaluations of identity, privacy, content, and safety unlearning. These benchmarks support standardized comparisons and provide complementary evidence for unlearning efficacy, utility retention, robustness, and efficiency across vision, language, audio, and generative settings.

\section{Evaluation Metrics Overview}

Evaluation of multimodal unlearning relies on metric suites that jointly characterize forgetting, utility retention, robustness, and efficiency, as summarized in Figure~\ref{fig:mmunlearn_eval_taxo}. Prior work measures forgetting using targeted performance drops and concept-suppression signals, and complements these with safety and privacy audits that probe refusal behavior and membership or identity leakage. Retained capability is then verified on non-forgotten data using task and generation quality metrics, while robustness and practicality are assessed via adversarial stress tests and compute or environmental budgets. We defer metric definitions and protocols to Appendix~\ref{app:evaluation_frameworks}, which consolidates formulations and validation procedures across vision, language, audio, and generative settings.

\begin{figure}[!h]
    \centering
    \includegraphics[width=1.0\columnwidth]{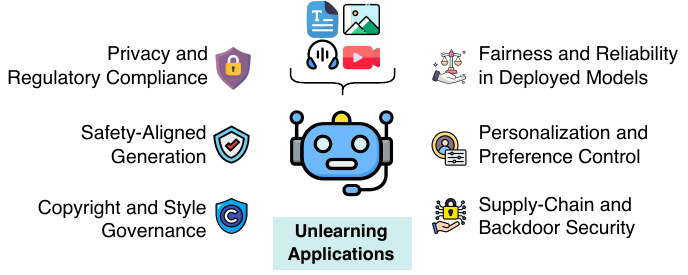}
    \caption{Core application scenarios of multimodal unlearning across privacy, safety, governance, personalization, and security. Details in App.~\ref{app:application}.}
    \label{fig:applications}
    \vspace{-12pt}
\end{figure}

\section{Multimodal Unlearning Applications}

Multimodal unlearning supports deployed settings that require selective removal of learned information without full retraining. Figure~\ref{fig:applications} summarizes the primary application scenarios. Although application settings differ in targets, constraints, and evaluation priorities, they share a common objective: remove specific identities, attributes, concepts, or behaviors while preserving general capability and stability. We defer detailed use cases and representative studies to Appendix~\ref{app:application}.



\begin{figure}[!h]
    \centering
    \includegraphics[width=0.80\columnwidth]{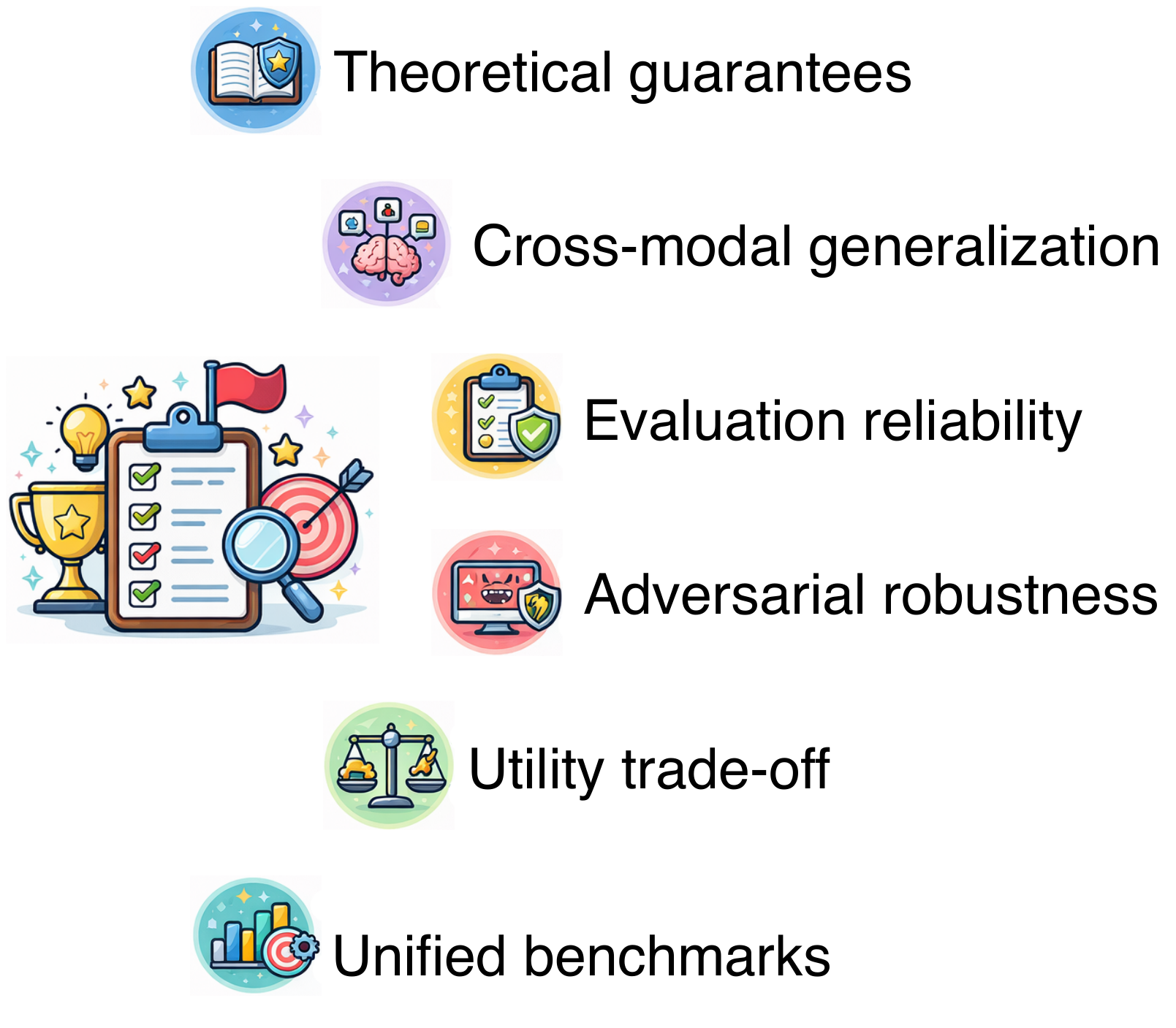}
    \caption{Key open challenges in multimodal unlearning across theory, generalization, evaluation, robustness, efficiency, and benchmarking. Details in App.~\ref{app:open_challenges}.}
    \label{fig:open_challenges}
    \vspace{-10pt}
\end{figure}

\section{Open Challenges}

Figure~\ref{fig:open_challenges} summarizes key open challenges in multimodal unlearning. We provide a more detailed discussion in Appendix~\ref{app:open_challenges}, covering modality-specific limitations, evaluation considerations, and emerging research directions, and highlighting open problems for reliable and scalable multimodal unlearning.

\section{Future Directions}

\textbf{Temporal and Dynamic Modalities.} Extending unlearning beyond static image-text pairs to temporal multimodal signals remains an open challenge. Existing work in audio and multimodal unlearning highlights the need to handle audio-vision coupling and speaker biometrics, raising unresolved questions around streaming, continual deletion, and deployment-time guarantees~\cite{liu2024multimodal, pathak2025quantum}. Parallel efforts in video and motion generation adapt unlearning to dynamic behaviors, including safety filtering and motion-aware personalization, but current methods remain limited in scope and evaluation~\cite{liu2024unlearning, de2025human}. 

\vspace{1pt}

\textbf{Frontier-Scale Model Unlearning.} Scaling unlearning methods and their evaluation to foundation-scale models remains an open challenge across modalities. Most existing studies operate on limited backbones, narrow concept scopes, or single-base architectures, which constrains conclusions about generalization to large multimodal foundation models~\cite{dontsov2024clear, patilunlearning2024, cheng2024mu}. 

\vspace{1pt}

\textbf{Sequential and Continual Unlearning.} Practical deployments require unlearning methods that remain effective under repeated deletions, downstream fine-tuning, and long update sequences. Recent work in multimodal LLMs highlights that performance and forgetting behavior can drift as deletions accumulate, motivating continual rather than one-shot unlearning protocols~\cite{kawakami2025pulse}. In generative diffusion models, studies show that forgotten concepts may resurface after subsequent training, prompting methods that aim to preserve deletion effects across sequential updates~\cite{suriyakumar2024unstable, li2025towards, li2025sculpting}. Designing unlearning mechanisms that remain stable under long-horizon updates therefore remains an open challenge.

\vspace{1pt}

\textbf{Controllable and Fine-Grained Unlearning.} Recent work increasingly targets fine-grained control over what is forgotten, shifting from coarse dataset-level deletion to data-point, attribute, and knowledge-unit unlearning in multimodal models and VLMs~\cite{li2024single, xing2024efuf, sinha2025multi}. Parallel efforts in speech, music, and diffusion models emphasize selective suppression of identity-, style-, or trigger-related features while preserving surrounding content and overall generation quality~\cite{cheng2025speech, kim2025no, liu2024efficient, park2024direct}. Across modalities, this setting exposes shared challenges in precision, compositionality, and stability under adversarial use or downstream adaptation, highlighting the need for unlearning mechanisms that provide reliable, interpretable, and scalable control across concepts and modalities~\cite{cywinski2025saeuron, zhang2024unlearncanvas}.

\vspace{1pt}

\textbf{Inference-Time Unlearning.} Inference-time mechanisms suppress undesired content during generation without modifying model parameters, offering reversible and deployment-friendly control. In text-to-image diffusion, guidance-path and conditioning-path controls adjust sampling trajectories or conditioning signals to steer generations away from unsafe or copyrighted concepts while keeping the base model fixed~\cite{li2024get, zhang2024steerdiff, han2025adaptive, park2025steering}. 

\vspace{1pt}

\textbf{Cross-Modal Leakage Mitigation.} Cross-modal leakage mitigation seeks to prevent unsafe, biased, or private information from transferring between modalities and to ensure consistent behavior across unimodal and multimodal settings. Prior studies show that safety or privacy alignment achieved in text does not reliably generalize to vision, audio, or joint reasoning, which motivates the development of multimodal attacks, metrics, and evaluation benchmarks that explicitly probe cross-modal leakage pathways~\cite{chakraborty2024can, patilunlearning2024, kawakami2025pulse, liu2025protecting}.

\section{Conclusion}
This survey presents a systematic review of multimodal unlearning as a core capability for accountable Multimodal Foundation Models (MFMs), with an emphasis on selective removal while preserving utility. By reviewing existing methods, highlighting emerging trends, and discussing open challenges, we adopt a system-oriented perspective that organizes unlearning mechanisms by intervention stage and control pathway, enabling comparison across vision, language, video, and audio models. Our synthesis highlights key gaps in evaluation reliability, robustness to adversarial reactivation, and deployment-facing constraints. Finally, we outline research directions toward unified benchmarks, stronger robustness guarantees, and tighter integration between unlearning mechanisms and deployment pipelines.

\section*{Limitations}
This survey aims to provide broad coverage of multimodal unlearning for foundation models, but several limitations remain. First, despite systematic efforts to include relevant studies published before submission, some recent or less visible works may be omitted due to the rapid pace of progress in this area. Second, the analysis prioritizes system-level perspectives, such as intervention stages and control pathways, rather than method-centric or algorithm optimization-oriented perspectives. Third, given the breadth of coverage, we do not detail algorithmic design and optimization and instead direct readers to the primary works that introduce these methods; presentation constraints limit deeper discussion of fine-grained taxonomies and modality-specific nuances, some of which are deferred to the appendix. In addition, as methods, datasets, and evaluation protocols evolve rapidly, maintaining a fully up-to-date taxonomy is challenging. However, our system-first taxonomy is designed to serve as a scaffold for organizing future developments. We hope this survey supports the continued development of multimodal unlearning in both academic and industrial settings. At the same time, several data types and settings remain beyond the present scope, including time series, tabular, sensor, and related structured or streaming data, among others, while audio and video unlearning remain comparatively underexplored.

\section*{Acknowledgments}
We thank Prof. Sijia Liu (Michigan State University) for helpful feedback and constructive suggestions. This work was supported in part by the Google PhD Fellowship.

\bibliography{custom}

\appendix

\newpage

\section{Additional Dataset Details}
\label{app:datasets}

Several specialized unlearning settings rely on targeted datasets to evaluate concept-level or domain-specific forgetting. Table~\ref{tab:person_copyright_datasets} covers personalization setup and copyright unlearning, as well as knowledge QA and instruction probes for factual or behavioral erasure in vision-language tasks, segmentation and image-to-image (I2I) unlearning for pixel-level concepts or stylistic attributes, and recommender unlearning for user-item interactions. Table~\ref{tab:speech_safety_datasets} summarizes datasets for speech unlearning (targeting speaker traits and linguistic content), safety robustness unlearning (evaluating resistance to jailbreak prompts and refusal consistency), and web-scale data hygiene, which focuses on removing noisy, sensitive, or undesirable patterns from large pretraining corpora. Finally, Table~\ref{tab:class_datasets} reports class unlearning benchmarks that evaluate the removal of entire semantic categories in classifiers using standard image datasets.

\section{Detailed Unlearning Evaluation Frameworks}
\label{app:evaluation_frameworks}

\subsection{Forget Quality and Safety}
\label{subsec:forget_quality}

\textbf{Unlearning Accuracy.} Unlearning Accuracy (UA) measures forgetting efficacy as the complement of predictive accuracy on the forget set~\cite{wu2025munba, schioppa2024model, sendera2025semu}:
\[
\mathrm{UA} = 100\% - \mathrm{Accuracy}(D_f),
\]
where $D_f$ denotes the subset designated for removal. Related forgetting-oriented metrics include Forget Accuracy, which reports post-unlearning accuracy on the forbidden class~\cite{pathak2025quantum}, and Removal Accuracy, which measures the fraction of attack triggers that no longer elicit the undesired behavior~\cite{aravindan2025sealing, jha2025backdoor}.

\textbf{Zero-Shot Forget Accuracy (FA@k).} For VLMs with zero-shot prediction, FA@k measures whether the true label of a forget example appears among the top-$k$ model predictions. Given a forget set $D_f$ and model scores $f(x)$,
\[
\mathrm{FA}@k
=
\frac{1}{\lvert D_f\rvert}
\sum_{(x,y)\in D_f}
\mathbf{1}\!\left\{\, y \in \operatorname{Top\mbox{-}k}\!\big(f(x)\big) \,\right\}.
\]
This metric is commonly reported for $k\in\{1,5\}$ in zero-shot VLM evaluations~\cite{caitargeted2025}.

\textbf{Degree of Unlearning.} Distributional change in concept scores before and after unlearning can be quantified using the 1-Wasserstein distance. Let $B$ denote the pre-unlearning score distribution, $A$ the post-unlearning distribution, and $R$ a reference distribution. The degree of unlearning is defined as
\[
\gamma = \frac{W_1(A,B)}{W_1(B,R)},
\]
where $W_1(\cdot,\cdot)$ denotes the 1-Wasserstein distance~\cite{solomon2014earth, tong2021diffusion}.

\vspace{1pt}

\textbf{CLIP Classification Drop.} Concept erasure in image generation can be verified through classification performance on generated samples. Let a generator produce $n$ images for a concept prompt before unlearning, $\{x_i^{\mathrm{pre}}\}_{i=1}^{n}$, and after unlearning, $\{x_i^{\mathrm{post}}\}_{i=1}^{n}$. Using a zero-shot CLIP classifier or a specialized detector $c(\cdot)\in\{0,1\}$, the classification drop is computed as
\[
\Delta_{\mathrm{cls}}=\frac{1}{n}\sum_{i=1}^{n} c\!\left(x_i^{\mathrm{pre}}\right)
-\frac{1}{n}\sum_{i=1}^{n} c\!\left(x_i^{\mathrm{post}}\right).
\]
A higher $\Delta_{\mathrm{cls}}$ indicates greater removal of the target concept from generated outputs. CLIP-based classification accuracy serves as a standard erasure indicator such as ESD~\cite{gandikota2023erasing}, MACE~\cite{lu2024mace}.

\vspace{1pt}

\textbf{CLIP Similarity Drop.} CLIP image-text similarity provides a continuous signal of residual concept alignment. Using the same image sets and the concept text $t$, let $f_{\mathrm{img}}, f_{\mathrm{text}}$ be CLIP encoders and let $\cos(\cdot,\cdot)$ denote cosine similarity. Define average similarities

\vspace{-10pt}

\[
s_{\mathrm{pre}} = \frac{1}{n}\sum_{i=1}^{n}\cos\!\big(f_{\mathrm{img}}(x_i^{\mathrm{pre}}),\, f_{\mathrm{text}}(t)\big)
\]
\vspace{-10pt}
\[
s_{\mathrm{post}} = \frac{1}{n}\sum_{i=1}^{n}\cos\!\big(f_{\mathrm{img}}(x_i^{\mathrm{post}}),\, f_{\mathrm{text}}(t)\big)
\]

and the similarity drop $\Delta_{\mathrm{sim}} = s_{\mathrm{pre}} - s_{\mathrm{post}}$. When $\Delta_{\mathrm{sim}}$ increases, alignment with the concept decreases. Empirical reports show that classifier confidence can collapse while CLIP similarity falls only slightly, so reporting both measures is helpful for diagnosing residual representations~\cite{gandikota2023erasing, rusanovsky2025memories, wang2025precise}.

\subsection{Safety \& Content Forgetting}
\label{subsec:safety_Content}

\textbf{Refusal Rate on Forbidden Prompts.} Also referred to as rejection rate, Refusal Rate (RR) measures how often the model refuses harmful queries after unlearning~\cite{chen2025safety}. Let $D$ be the evaluation set of harmful text-image inputs and $R_i$ the model response to the $i$-th prompt. Define the refusal indicator $I_{\mathrm{ref}}(R_i)=1$ if the response contains refusal content (per a predefined policy template) and $0$ otherwise. The metric is
\[
RR = \frac{1}{\lvert D\rvert} \sum_{i=1}^{\lvert D\rvert} I_R\!\left(R_i\right),
\]
so higher RR indicates more consistent rejection of harmful requests. 

\vspace{1pt}

\textbf{Inappropriate Content Rate.} This metric measures how often a model produces unsafe content under sensitive prompts. In image generation, a standard protocol samples outputs and reports the fraction flagged by external NSFW detectors (e.g., Q16 or NudeNet), where lower post-unlearning rates indicate safer behavior~\cite{schramowski2023safe}. Let $Y_{\mathrm{pre}}=\{y_i^{\mathrm{pre}}\}_{i=1}^{n}$ and $Y_{\mathrm{post}}=\{y_i^{\mathrm{post}}\}_{i=1}^{n}$ denote outputs before and after unlearning for the same prompt set, and let $IR_{\mathrm{pre}}$ and $IR_{\mathrm{post}}$ be the corresponding flagged fractions under a binary detector $d(\cdot)\in\{0,1\}$. The improvement is summarized by the drop $\Delta IR = IR_{\mathrm{pre}} - IR_{\mathrm{post}}$. Several works also estimate harm with an LLM-based judge (optionally via image captions) and aggregate scores by thresholding or averaging~\cite{wang2024moderator}.

\vspace{1pt}

\textbf{VLM‑Based Judgments.} Pretrained VLMs can serve as external judges for presence of a forbidden concept. Let a VQA-style judge output a binary decision $g(y)\in\{0,1\}$ for concept presence, or a matching score $s(y,t)\in[0,1]$ for image $y$ and concept text $t$. Define the yes-rate drop and similarity drop as

\vspace{-20pt}

\[
\Delta_{\mathrm{VQA}} = \frac{1}{n}\sum_{i=1}^{n} g(y_i^{\mathrm{pre}}) -
\frac{1}{n}\sum_{i=1}^{n} g(y_i^{\mathrm{post}}),
\]
\[
\Delta_{s} = \frac{1}{n}\sum_{i=1}^{n} s(y_i^{\mathrm{pre}},t) -
\frac{1}{n}\sum_{i=1}^{n} s(y_i^{\mathrm{post}},t).
\]
VLMs used for $g$ or $s$ include VQA heads such as CLIP-FlanT5-based VQAScore and ITM scores from BLIP-2; these are standard tools for judging whether generated content still expresses the concept~\cite{lin2024evaluating}. Larger $\Delta_{\mathrm{VQA}}$ or $\Delta_{s}$ indicates more effective forgetting.

\subsection{Attack-Based Privacy}
\label{subsec:attack_based_privacy}

\textbf{Membership Inference Attack and Enhanced Variants.}
Membership Inference Attacks (MIA) are a standard privacy test for evaluating whether an unlearned model still leaks information about forgotten data. MIA estimates how easily an adversary can infer whether a sample was part of the original training set. For a forget set $D_f$, following established formulations~\cite{shokri2017membership, carlini2022membership, jia2023model, wang2025membership}, MIA efficacy is defined as
\[
\mathrm{MIA}
=
\frac{1}{\lvert D_f \rvert}
\sum_{x_i \in D_f}
\mathbf{1}\!\big[\,A(F_T, x_i) \in \{0,1\}\,\big],
\]
where $F_T$ denotes the evaluated target model and $A$ the membership inference attacker, which predicts membership as $1$ if $x_i \in D_{\mathrm{train}}$ and $0$ otherwise. Higher MIA efficacy indicates that the unlearned model behaves closer to a model retrained without the forgotten data. Beyond the basic setting, prior work proposes enhanced MIA variants that audit specific components or compare unlearned models against retrained references, providing stronger privacy guarantees~\cite{dontsov2024clear, wang2025muc, koudounas2025alexa}. 

\vspace{1pt}

\textbf{Identity Matching.} Identity leakage metrics assess whether model outputs still reveal a forgotten identity after unlearning. In vision settings, evaluation typically relies on recognition accuracy or embedding similarity between generated outputs and reference images. Forgetting is considered successful when recognition accuracy for the erased identity drops to chance level and embedding similarity exhibits a substantial decline~\cite{biswas2025cure, caitargeted2025, nagasubramaniam2025prompting}. Common embedding-based measures include Identity Matching Score (IMS)~\cite{liu2024metacloak} and Identity Score Matching (ISM)~\cite{wu2025unlearning}. In text and multimodal evaluations, identity leakage is monitored through identity mentions in generated captions or VQA responses, where effective erasure drives correct mention rates toward zero~\cite{dontsov2024clear, ma2024benchmarking}. 

\vspace{1pt}

\textbf{Voice Privacy.} In speech unlearning, privacy evaluation assesses whether a model can still recognize or reproduce a forgotten speaker after unlearning. A common signal is speaker similarity (SIM), which measures the alignment between embeddings of generated and reference utterances; effective unlearning reduces SIM for forgotten speakers while preserving similarity for retained ones~\cite{chen2022wavlm}. 

Complementary to similarity, speaker Zero-Retrain Forgetting (spk-ZRF)~\cite{kim2025not} evaluates whether speaker identity becomes uncorrelated with prompting after unlearning. It computes the Jensen-Shannon divergence between speaker identity distributions obtained with and without speaker prompts,
\[
JSD_i = \frac{1}{2}\left[
D_{\mathrm{KL}}\!\left(p_i \,\|\, m_i\right)
+
D_{\mathrm{KL}}\!\left(q_i \,\|\, m_i\right)
\right]
\]

\vspace{-10pt}

\[
\mathrm{spk\text{-}ZRF} =
1 -
\frac{1}{n_f}
\sum_{i=1}^{n_f} JSD_i,
\]
where higher spk-ZRF values indicate that generated speech no longer preserves the forgotten speaker identity.

\subsection{Model Utility and Faithfulness} 
\label{subsec:model_utility}

\textbf{Classification Accuracy.}
Retained utility on non-forgotten data is commonly measured by Top-$k$ classification accuracy on remaining classes~\cite{bansal2023cleanclip, struppek2024exploiting, han2025unlearning, biswas2025cure}:
\[
\mathrm{Top\mbox{-}k\ Acc}
=
\frac{1}{N}\sum_{i=1}^{N}\mathbf{1}\!\left[\,y_i \in \mathrm{Top\mbox{-}k}\big(\hat{\mathbf{p}}_i\big)\,\right],
\]
where $y_i$ denotes the ground-truth label and $\hat{\mathbf{p}}_i$ the predicted class scores.

\textbf{Cross-Modal Retrieval Utility.}
For multimodal models, utility retention is evaluated using retrieval metrics such as Recall@$K$ and R-Precision on held-out benchmarks~\cite{yang2024cliperase, sinha2025multi}:
\[
\mathrm{Recall@}K
=
\frac{1}{N}\sum_{i=1}^{N}
\mathbf{1}\!\left[\,R(q_i)\cap \mathrm{TopK}(q_i)\neq\varnothing\,\right].
\]

\textbf{Language and QA Metrics.} Retained capability on non-forgotten data is tracked with standard NLP scores. For VLMs that perform question answering or caption generation, language quality on non-forgotten examples is assessed with BLEU~\cite{zhang2025does}, ROUGE-L~\cite{dontsov2024clear}, and METEOR~\cite{liu2024large}. Stable BLEU/ROUGE-L/METEOR on unrelated VQA or captioning items indicates preserved language utility. In addition, CLIP Score~\cite{hessel2021clipscore} is widely used to assess image-text alignment, with consistent scores on non-target prompts suggesting that multimodal semantic alignment remains intact following unlearning~\cite{yang2024cliperase, cheng2024multidelete}.

\vspace{1pt}

\textbf{Generative Output Quality.} To ensure image generation quality is retained, vision metrics like Fréchet Inception Distance (FID)~\cite{heusel2017gans}, Fréchet Video Distance (FVD)~\cite{unterthiner2019fvd, facchiano2025video}, Kernel Inception Distance (KID)~\cite{binkowski2018demystifying} and inverted FID (IFID)~\cite{li2024get} are commonly reported. These metrics compare the distribution of generated images to that of real images using feature statistics. FID computes the distance between the means $(\mu)$ and covariances $(\Sigma)$ of Inception features for generated $(g)$ and real $(r)$ samples:
\[
\begin{aligned}
\mathrm{FID}(r,g)
&= \lVert \mu_r - \mu_g \rVert_2^2 \\
&\quad + \operatorname{Tr}\!\big(
\Sigma_r + \Sigma_g - 2(\Sigma_r \Sigma_g)^{1/2}
\big).
\end{aligned}
\]
Lower FID and stable KID values on retain-set prompts indicate that unlearning preserves fidelity and diversity of generated images~\cite{fan2023salun, zhang2024unlearncanvas, chen2025score}. 

Beyond distributional similarity, perceptual and faithfulness metrics provide complementary signals. PickScore~\cite{kirstain2023pick} and Aesthetic Score (AES)~\cite{schuhmann2022laion} evaluate semantic alignment and visual appeal, while Polling-based Object Probing Evaluation (POPE)~\cite{li2023evaluating} measures object hallucination in VLM outputs; stable scores suggest that unlearning does not degrade perceptual quality or semantic correctness~\cite{ma2024benchmarking, li2024single}.

\textbf{Perceptual Similarity.} Perceptual similarity metrics assess whether unlearning alters model outputs on benign inputs by comparing generations from the unlearned model to those of the original model. The Learned Perceptual Image Patch Similarity (LPIPS) score~\cite{zhang2018unreasonable} measures perceptual distance between two images in a deep feature space. Lower LPIPS values on retain prompts indicate higher integrity, meaning that outputs remain perceptually close on non-target inputs after unlearning. Mean LPIPS on benign prompts is therefore commonly reported to verify that unlearning preserves visual details, style, and overall generation quality~\cite{dai2023training, park2024direct, biswas2025cure}.

\textbf{LLM-as-a-Judge Evaluation.} Several multimodal unlearning studies use large language models as semantic evaluators to score model outputs. These approaches prompt an LLM with task-specific rubrics and interpret its responses as scores for safety, factuality, or answer quality. Recent multimodal benchmarks adopt GPT-Eval-style setups to rate generated outputs along these semantic dimensions~\cite{ma2024benchmarking, park2024direct, xu2025pebench, liu2025protecting}. Such evaluations provide a semantics-aware assessment of unlearning behavior that complements surface-level automatic metrics.

\textbf{Human-Centered Evaluation.} While most unlearning work relies on automatic metrics, several multimodal studies incorporate human judgment to assess perceived safety and fidelity. In safety-oriented evaluations, annotators label model outputs from different training or unlearning conditions for harmfulness, and aggregated judgments with high inter-annotator agreement reveal changes in harmful output rates after unlearning~\cite{chakraborty2024can}. In diffusion unlearning, human studies compare generated images against reference subjects to assess whether unlearning suppresses identity- or style-specific resemblance while preserving benign generations~\cite{huang2024freezeasguard}. These evaluations provide complementary evidence that unlearning reduces harmful or identifiable content beyond what automated metrics capture.

\subsection{Adversarial Perturbation Robustness} 
\label{subsec:adversarial_perturbation}

Attack Success Rate (ASR) quantifies how often adversarially perturbed inputs still elicit forbidden content from an unlearned model. Let $D$ be the evaluation set of harmful text-image pairs and $R_i = f(x_i^{\mathrm{adv}})$ the response to the $i$-th adversarial input; a response is unsafe if it contains forbidden content. The ASR is defined as
\[
\mathrm{ASR}
=
\frac{1}{\lvert D\rvert}
\sum_{i=1}^{\lvert D\rvert}
I_A\!\left(R_i\right),
\]
where $I_A(\cdot)$ is an indicator that returns $1$ when the response contains harmful knowledge and $0$ otherwise~\cite{chen2025safeeraser}. A higher ASR indicates that forgotten content remains vulnerable to adversarial reactivation, suggesting incomplete unlearning. Prior work reports ASR under both white-box and black-box attack settings to assess robustness of unlearning against adaptive adversaries~\cite{bansal2023cleanclip, zhang2024steerdiff, biswas2025cure}.

\subsection{Compute and Environmental Budget}
\label{subsec:compute_budget}

\textbf{Run-time and Memory Usage.} Compute footprint anchors the edit budget for unlearning methods. Studies now report wall-clock runtime (often denoted WCT) and peak memory as first-class metrics under Run-Time Efficiency (RTE, typically measured in minutes), alongside peak GPU memory consumption (in GB), to certify that forgetting is practical at scale. Beyond elapsed time, some work also quantifies training cost using total floating-point operations (TFLOPs) and effective throughput (TFLOPS), and characterises inference cost via a relative complexity ratio with respect to a backbone model~\cite{zhang2024learning}. Across image classification, diffusion, and contrastive settings, recent work consistently reports WCT, memory usage, and FLOP-based measures, showing modest additional compute compared to full retraining and making unlearning overheads comparable across architectures and hardware platforms~\cite{fan2023salun, li2025loreun, dang2025diffzoo, cywinski2025saeuron, wang2025muc, spartalis2025unleashing}.

\textbf{Environmental Cost.} Beyond accuracy and robustness, multimodal unlearning also introduces an environmental cost. Recent work estimates emissions by logging GPU energy in kilowatt-hours and multiplying by an assumed grid carbon intensity of about 0.4 kgCO$_2$e per kWh~\cite{dodge2022measuring, chakraborty2024can}. These measurements show that multimodal unlearning consumes substantially more energy than text-only unlearning on the same GPU, so reporting energy use and derived CO$_2$e for each setting helps evaluations of unlearning account for environmental impact alongside safety and privacy.

\section{Unlearning Robustness}

\textbf{Adversarial Reactivation Attacks.} Adversarial reactivation attacks evaluate unlearning robustness by optimizing prompts or guidance that recover a forgotten concept without modifying model weights. These attacks exploit residual conditioning, safety, or cross-modal pathways and operate at decoding or prompting time using gradient-based, surrogate, or zeroth-order search~\cite{kim2024automatic, dang2025diffzoo, zhang2025does}.

\vspace{-10pt}

\[
\begin{aligned}
\max_{p,z}\;\;& S_{\theta}(p,z;c)\;-\;\lambda_{1}\,\Delta(p,p_{0})\;-\;\lambda_{2}\,R(z)\\
\text{s.t. }\;\;& \text{queries}\le Q_{\max},\quad C(p)\in \mathcal{B}.
\end{aligned}
\]

Here \(\theta\) denotes fixed model parameters; \(p\) is a discrete prompt and \(z\) an optional conditioning latent or embedding; \(S_{\theta}(p,z;c)\) scores concept \(c\) (for example CLIP similarity, an NSFW detector logit, or a task success score); \(\Delta\) bounds prompt edits from a seed \(p_{0}\); \(R\) regularizes latents; \(\mathcal{B}\) enforces benign surface form and \(Q_{\max}\) limits black-box queries. Transfer terms or surrogate models can be included by adding \(\alpha\,\mathbb{E}_{\phi}\,S_{\phi}(p,z;c)\) to encourage cross-model success~\cite{han2024probing, liu2025image}.

Methods differ in how they optimize this objective. AutoJailbreaking~\cite{kim2024automatic}, which performs LLM-driven prompt search to evade filters and reveal residual unsafe behavior; DiffZOO~\cite{dang2025diffzoo}, which uses query efficient zeroth order ascent in the discrete token space to elicit the target under strict black box budgets; and Stealthy MLLM~\cite{zhang2025does}, which designs distribution shifted or dual purpose prompts that pass standard checks yet recover forgotten answers, exposing evaluation blind spots.

\vspace{3pt}

\textbf{Inference‑time Defenses.} Inference-time defenses mitigate residual failures after unlearning by intervening during sampling rather than modifying parameters. They operate on the conditioning stream to suppress adversarial signals while preserving responses to benign prompts, commonly through subspace projection of adversarial token directions or adaptive smoothing of token activations~\cite{chen2025dual, han2025adaptive}.

\vspace{-15pt}

\[
s^{\mathrm{def}}_{t}(x_t, E)
= s_{\theta}\!\bigl(x_t \mid S_t(\Pi_{\perp} E)\bigr),
\Pi_{\perp} = I - U U^{\top},
\]

Here $x_t$ denotes the latent at timestep $t$, $E$ the matrix of text token embeddings, and $s_\theta$ the conditional score function. The matrix $U$ spans an estimated adversarial subspace, and $\Pi_{\perp}$ projects embeddings orthogonally to that subspace. The operator $S_t$ applies token-wise smoothing, such as median filtering, before scoring. Setting $S_t$ to the identity recovers pure projection, while setting $U$ to zero recovers adaptive smoothing.

\section{Unlearning‑Adjacent Controls}

\textbf{Bias and Privacy Safeguards.} Bias and privacy safeguards intervene on the data path. They constrain what the model sees and how prompts are encoded before any weight update, so optimization proceeds on balanced evidence with reduced attribute leakage~\cite{alabdulmohsinclip2024, huang2024enhancing, liu2024pre}.

\vspace{-15pt}

\[
\begin{aligned}
\min_{\theta}\;& \mathbb{E}_{(x,y)\sim D}\big[w_{\mathrm{bal}}(y)\,L(f_{\theta}(x),y)\big]\\
&+ \lambda\,R_{\mathrm{priv}}(f_{\theta};A,g),
\end{aligned}
\]

where $f_\theta$ is the model, $L$ the task loss, $w_{\mathrm{bal}}(y)$ denotes class- or attribute-level reweighting for bias control, $A$ indexes sensitive attributes, $g$ is a privacy editing operator such as differentially private image sanitization, and $R_{\mathrm{priv}}$ penalizes residual attribute leakage.

\textbf{In-Context Mitigation.} In-context mitigation steers a frozen VLM at prompt time by inserting a small set of curated multimodal demonstrations and summaries, so that decoding conditions on safer evidence rather than on harmful patterns~\cite{zhou2024visual}. Because it operates entirely through the input channel, it avoids retraining and remains reversible, but its effectiveness depends on demonstration quality, retrieval coverage, and the available context budget.
\begin{table*}[t!]
    \centering\small
    \begin{tabular}{p{0.10\textwidth} p{0.30\textwidth} p{0.24\textwidth} p{0.26\textwidth}}
        \toprule
        \textbf{Modality} & \textbf{Dataset} & \textbf{Size} & \textbf{Used in} \\

        \midrule
        \rowcolor{gray!25} 
        \multicolumn{4}{c}{%
          \rule{0pt}{2.2ex}\textbf{Personalization Setup}\rule[-.9ex]{0pt}{0pt}
        } \\
        \midrule
        
        \multirow{1}{*}{Image}
            & DreamBooth~\cite{ruiz2023dreambooth}        & 30 subjects, 4-6 images each & \citealp{liu2024metacloak, li2025towards} \\  
        \midrule

        \multirow{2}{*}{Image-Text}
            & DiffusionDB~\cite{wang2023diffusiondb}             & 14M images, 1.8M prompts & \citealp{pan2024leveraging, li2025towards} \\
            \cmidrule(lr){2-4}

            & DreamBench++~\cite{pengdreambench2025}          & 150 images with 1,350 prompts & \citealp{li2025towards} \\
        \midrule

        \rowcolor{gray!25} 
        \multicolumn{4}{c}{%
          \rule{0pt}{2.2ex}\textbf{Copyright Unlearning}\rule[-.9ex]{0pt}{0pt}
        } \\
        \midrule

        \multirow{2}{*}{Image}
            & CPDM~\cite{ma2024dataset}       & 2.1K anchors and 18.9K paired generated images & \citealp{moon2024holistic, liu2025rethinking, jin2025unconsciously, ren2025six} \\
            \cmidrule(lr){2-4}
            
            & VioT~\cite{kim2024automatic}        & 100 images total across 5 copyrighted categories & \citealp{kim2024automatic} \\
            
        \midrule

        \multirow{1}{*}{Audio}
            
            & MusicCaps~\cite{agostinelli2023musiclm}       & 5.5K captioned clips & \citealp{kim2025no} \\
        \midrule

        \rowcolor{gray!25} 
        \multicolumn{4}{c}{%
          \rule{0pt}{2.2ex}\textbf{Knowledge QA and Instruction Probes}\rule[-.9ex]{0pt}{0pt}
        } \\
        \midrule

        \multirow{10}{*}{Image-Text}
            & VQA~\cite{antol2015vqa} & 255K images, 764K questions, 10M human answers & \citealp{ma2024benchmarking, dontsov2024clear, chen2025safety} \\
            \cmidrule(lr){2-4}

            & VQAv2~\cite{goyal2017making} &  265K images with 1.1M questions & \citealp{li2024single, chakraborty2024can, chen2025safety} \\
            \cmidrule(lr){2-4}
            
            & NLVR2~\cite{suhr2019corpus}        & 107K caption-image pairs, 29.7K unique sentences & \citealp{cheng2024mu, cheng2024multidelete} \\
            \cmidrule(lr){2-4}
            
            
            & ScienceQA~\cite{lu2022learn}         & 21.2K multimodal multiple-choice science questions & \citealp{gao2024large, chen2025safety} \\
            \cmidrule(lr){2-4}

            & GQA~\cite{hudson2019gqa}        & 113K images with 22.7M compositional visual questions & \citealp{li2024single, xing2024efuf, li2024get} \\
            \cmidrule(lr){2-4}


            & UnLOK‑VQA~\cite{patilunlearning2024} & 500 visual QA samples (OK-VQA~\cite{marino2019ok} extension) & \citealp{patilunlearning2024, wu2025medforget} \\
            \cmidrule(lr){2-4}

            & VizWiz~\cite{gurari2018vizwiz} & 31K real-world visual questions from blind users & \citealp{li2024single, chen2025safety, chen2025safeeraser} \\
            \cmidrule(lr){2-4}
            

            & POPE~\cite{li2023evaluating} & 18K object-image queries for VLM hallucination evaluation & \citealp{xing2024efuf, li2024single, ma2024benchmarking, xu2025pebench} \\

        \midrule
        
        \multirow{1}{*}{Text}
            & PGR~\cite{sousa2019silver}     & 1.7K PubMed abstracts annotated with 4.2K phenotype-gene relations & \citealp{cheng2024multidelete}  \\
        \midrule
        
        \rowcolor{gray!25} 
        \multicolumn{4}{c}{%
          \rule{0pt}{2.2ex}\textbf{Segmentation and I2I Unlearning}\rule[-.9ex]{0pt}{0pt}
        } \\
        \midrule

        \multirow{2}{*}{Image}
            & MS‑COCO~\cite{lin2014microsoft}     & 2.5M labeled instances in 328K images (80 classes) & \citealp{park2024direct, xing2024efuf, cywinski2025saeuron, polowczyk2025unguide} \\
            \cmidrule(lr){2-4}

            & UnlearnCanvas~\cite{zhang2024unlearncanvas} & 60 artistic styles across 20 object categories & \citealp{caitargeted2025, zhang2025concept, cywinski2025saeuron} \\
            
        \midrule
        
        \rowcolor{gray!25} 
        \multicolumn{4}{c}{%
          \rule{0pt}{2.2ex}\textbf{Recommender Unlearning}\rule[-.9ex]{0pt}{0pt}
        } \\
        \midrule

        \multirow{1}{*}{Image-Text}
            & Amazon Reviews~\cite{hou2024bridging} & 571.5M reviews from 54.5 M users on 48.2 M items across 33 categories & \citealp{sinha2025multi} \\
        \midrule

        \multirow{1}{*}{Text-Graph}
            & Amazon Products~\cite{hou2024bridging} & 9.3M items, 144M reviews, 237M relational edges & \citealp{dang2025efficient} \\

            
        \midrule

        \multirow{1}{*}{Text-Metadata}
            & Yelp~\cite{yelp_open_dataset} & 6.9M reviews, 150K businesses, with user, check-in, tip, and photo data & \citealp{dang2025efficient} \\


            
            
            
        \bottomrule
        
    \end{tabular}
    \caption{Datasets are grouped by unlearning setting (Personalization Setup; Copyright Unlearning; Knowledge QA and Instruction Probes; Segmentation and I2I Unlearning; Recommender Unlearning) and modality, with their sizes and representative studies.}
    \label{tab:person_copyright_datasets}
\end{table*}


\begin{table*}[t!]
    \centering\small
    \begin{tabular}{p{0.08\textwidth} p{0.28\textwidth} p{0.24\textwidth} p{0.26\textwidth}}
        \toprule
        \textbf{Modality} & \textbf{Dataset} & \textbf{Size} & \textbf{Used in} \\

        \midrule
        \rowcolor{gray!25} 
        \multicolumn{4}{c}{%
          \rule{0pt}{2.2ex}\textbf{Speech Unlearning}\rule[-.9ex]{0pt}{0pt}
        } \\
        \midrule

        \multirow{2}{*}{Audio}
            & Speech Commands~\cite{warden2018speech} & 64.7K v1 (30words, 1.9K speakers) /105.8K v2 (35words, 2.6K speakers) utterances & \citealp{cheng2024mu, cheng2025speech, pathak2025quantum} \\
            \cmidrule(lr){2-4}
            
            & AudioMNIST~\cite{becker2024audiomnist} & 30K spoken-digit (0–9) audio samples from 60 speakers (9.5 hours total) & \citealp{pathak2025quantum, mason-williams2025machine} \\
        \midrule
            
        \multirow{2}{*}{Audio-Text}
            & LibriSpeech~\cite{panayotov2015librispeech}      & 1,000 h read English speech from 2.5K speakers, with transcripts  & \citealp{kim2025not, pathak2025quantum, liu2025unlearning} \\
            \cmidrule(lr){2-4}
            
            
            


            & ITALIC~\cite{koudounas2023italic} & 16.5K Italian intent audio samples (15.5 h), 70 speakers, 18 domains, 60 intents & \citealp{koudounas2025alexa} \\

            
        \midrule

        \rowcolor{gray!25} 
        \multicolumn{4}{c}{%
          \rule{0pt}{2.2ex}\textbf{Safety Robustness Unlearning}\rule[-.9ex]{0pt}{0pt}
        } \\
        \midrule

        \multirow{5}{*}{Image-Text}
            & I2P~\cite{schramowski2023safe} & 4.7K text-to-image prompts for inappropriate-content evaluation & \citealp{fan2023salun, park2024direct, wu2025munba, moon2024holistic, ko2024boosting, cywinski2025saeuron, li2025loreun, li2025detect, li2025sculpting}\\
            \cmidrule(lr){2-4}

            & SneakyPrompt / NSFW\_200~\cite{yang2024sneakyprompt} & 200 NSFW prompts and 100 dog/cat scenario prompts & \citealp{li2024safegen, wang2024moderator, park2024direct, zhang2024steerdiff}\\
            \cmidrule(lr){2-4}

            
            & NudeNet~\cite{bedapudi2019nudenet} & 160K training images (auto-labeled) for nudity detection ($>$700K web-scraped images) & \citealp{poppi2023removing, han2024probing, shirkavand2025efficient, dang2025diffzoo, chen2025dual}\\
            \cmidrule(lr){2-4}

            & MIS~\cite{ding2025rethinking} & 6.2K multi-image safety samples & \citealp{chen2025safety, hu2025vlsbench} \\
            \cmidrule(lr){2-4}
            
            & FigStep~\cite{gong2025figstep} & 500 harmful questions over 10 safety topics & \citealp{chakraborty2024can, chen2025safety, zhang2025sua, chen2025safeeraser} \\

        \midrule

        \multirow{1}{*}{Video-Text}
            & SafeSora~\cite{dai2024safesora} & 14.7K prompts, 57.3K videos, 51.7K human safety annotations & \citealp{yoon2025safree, xu2025videoeraser} \\
        \midrule

        \rowcolor{gray!25} 
        \multicolumn{4}{c}{%
          \rule{0pt}{2.2ex}\textbf{Web-Scale Data Hygiene via Unlearning}\rule[-.9ex]{0pt}{0pt}
        } \\
        \midrule

        \multirow{3}{*}{Image-Text}
            & LAION-400M~\cite{schuhmann2021laion}        & 400M CLIP-filtered image-text pairs & \citealp{poppi2023removing, caitargeted2025} \\
            \cmidrule(lr){2-4}
            
            & CC3M~\cite{sharma2018conceptual}         & 3.3M web-harvested image-caption pairs & \citealp{bansal2023cleanclip, liang2024badclip, han2025unlearning}\\
            \cmidrule(lr){2-4}
            
            & Flickr30K~\cite{young2014image} & 31K images with 158K captions & \citealp{alabdulmohsinclip2024, liu2024multimodal, han2025unlearning} \\
            
        \bottomrule
        
    \end{tabular}
    \caption{Datasets are grouped by unlearning setting (Speech Unlearning; Safety Robustness Unlearning; Web-Scale Data Hygiene via Unlearning) and modality, with their sizes and representative studies.}
    \label{tab:speech_safety_datasets}
\end{table*}

\begin{table*}[t!]
    \centering\small
    \begin{tabular}{p{0.08\textwidth} p{0.30\textwidth} p{0.24\textwidth} p{0.26\textwidth}}
        \toprule
        \textbf{Modality} & \textbf{Dataset} & \textbf{Size} & \textbf{Used in} \\
        \midrule

        \rowcolor{gray!25} 
        \multicolumn{4}{c}{%
          \rule{0pt}{2.2ex}\textbf{Class Unlearning}\rule[-.9ex]{0pt}{0pt}
        } \\
        \midrule

        \multirow{12}{*}{Image}
            & ImageNet~\cite{deng2009imagenet}             & 3.2M images across 5.2K categories (synsets) & \citealp{zhang2023unlearnable, fan2023salun, han2025unlearning, caitargeted2025}\\
            \cmidrule(lr){2-4}
            
            & CIFAR~\cite{krizhevsky2009learning}         & 60K images; 10 classes (CIFAR-10) or 100 classes (CIFAR-100) & \citealp{fan2023salun, kim2024negmerge, ko2024boosting, sendera2025semu}\\
            \cmidrule(lr){2-4}
            
            & MNIST~\cite{lecun2002gradient}    & 70K grayscale handwritten digit images & \citealp{zhou2024visual, alberti2025data}\\
            \cmidrule(lr){2-4}
            
            & SVHN~\cite{netzer2011reading}    & 600K digit images from Street View (10 classes) & \citealp{fan2023salun, kim2024negmerge, wu2025munba}\\
            \cmidrule(lr){2-4}
            
            & Imagenette~\cite{Howard_Imagenette_2019}    & 13K images across 10 ImageNet classes & \citealp{fan2023salun, bui2025hiding, wu2025munba, biswas2025cure}\\
            \cmidrule(lr){2-4}

            & Stanford Cars~\cite{krause20133d}        & 16K images of 196 car classes & \citealp{zhang2023unlearnable, alabdulmohsinclip2024} \\
            \cmidrule(lr){2-4}

            & Stanford Dogs~\cite{khosla2011novel}        & 20K images of 120 dog breeds & \citealp{kravets2025zero, kravets2025clip} \\
            \cmidrule(lr){2-4}

            & Food-101~\cite{bossard2014food}        & 101K food images across 101 cuisine classes & \citealp{zhang2023unlearnable, liu2024efficient, han2025unlearning} \\
            \cmidrule(lr){2-4}

            
            & DTD~\cite{cimpoi2014describing}        & 5.6K texture images covering 47 describable categories & \citealp{ilharcoediting2023, alabdulmohsinclip2024} \\
            \cmidrule(lr){2-4}

            & SUN397~\cite{xiao2016sun}        & 108.7K images, 397 scene classes & \citealp{ zhang2023unlearnable, kim2024negmerge, han2025unlearning} \\
            \cmidrule(lr){2-4}

            & WikiArt~\cite{saleh2015large}        & 81K artwork images across 27 styles and 45 genres & \citealp{ma2024dataset, biggs2024diffusion, chen2025dual} \\
            
        \bottomrule
        
    \end{tabular}
    \caption{Datasets are grouped by unlearning setting (Class Unlearning) and modality, with their sizes and representative studies.}
    \label{tab:class_datasets}
\end{table*}


\section{Comprehensive Application Scenarios}
\label{app:application}

\textbf{Privacy and Regulatory Compliance.}
Unlearning for privacy and regulatory compliance addresses deletion requests, right-to-be-forgotten (RTBF) enforcement, and license-driven removals across multimodal systems. In vision-language pipelines, unlearning is used to erase specific identities, sensitive attributes, or marked image-text pairs while preserving general utility. Representative studies focus on identity- and pair-level deletion, supported by auditing datasets and evaluations that verify the suppression of sensitive answers or visual traits~\cite{cheng2024multidelete, dontsov2024clear, ma2024benchmarking}. In generative settings, diffusion-based work further formalizes compliant data removal within image generation pipelines~\cite{li2024machine}.

This application setting also includes consent-oriented and preventive controls that regulate how personal data enters training pipelines. Data-side protection mechanisms, such as unlearnable examples, introduce perturbations that prevent models from learning from protected samples, allowing individuals to share images or image-text pairs that resist downstream training while leaving unprotected data usable~\cite{zhang2023unlearnable}. Related ideas extend to structured perception tasks, providing model-agnostic protection across training pipelines~\cite{sun2024unseg}. Interactive privacy frameworks further integrate these capabilities by enabling contributors to control reuse of personal identities or styles and to request redaction or deletion through user-applied perturbations coupled with generative models and unlearning~\cite{liu2024metacloak}. Beyond vision, privacy-driven unlearning extends to speech and audio systems, where it supports speaker opt-out and private utterance deletion to meet RTBF-style requirements. Prior work demonstrates speaker-level forgetting and compliance-oriented evaluation in speech generation and recognition frameworks~\cite{kim2025not, cheng2025speech}.

\vspace{1pt}

\textbf{Safety-Aligned Generation.} Safety-aligned generation applies unlearning to remove NSFW, harmful, or toxic content while preserving benign behavior across modalities. In LLMs and VLMs, unlearning functions as a targeted safety control that suppresses unsafe behaviors without degrading general question answering or captioning performance~\cite{chakraborty2024can, chen2025safeeraser}. For VLMs, removing unsafe associations from cross-modal encoders yields safer retrieval and generation behavior under downstream use~\cite{poppi2023removing}.

In generative models, diffusion-based unlearning suppresses harmful visual concepts while maintaining output diversity and quality~\cite{fan2023salun, li2024safegen}. Similar safety-oriented edits extend to video and motion generation, where unlearning reduces unsafe or restricted content while preserving temporal coherence and realism~\cite{liu2024unlearning, de2025human}.

\vspace{1pt}

\textbf{Copyright and Style Governance.} Copyright and style governance in generative models leverages unlearning to remove protected styles or copyrighted content and to evaluate the completeness of such removal. In text-image diffusion, concept-level editing supports takedown of protected styles or instances, while benchmark datasets and standardized metrics assess whether copyrighted or identity-linked content has been effectively erased under copyright-sensitive deployments~\cite{kumari2023ablating, ma2024dataset, biswas2025cure}. Beyond still images, unlearning extends to other generative modalities. Prior work explores opt-out unlearning in music generation and applies concept-level removal in text-to-video diffusion to suppress copyrighted or IP-restricted content while preserving general generation quality~\cite{kim2025no, liu2024unlearning}.

\vspace{1pt}

\textbf{Fairness and Reliability in Deployed Models.} Fairness and reliability considerations motivate unlearning in deployed multimodal systems to mitigate biased, noisy, or unstable associations while preserving general capability. Fairness-oriented work leverages targeted forgetting to reduce skewed or culturally imbalanced associations in VLMs~\cite{struppek2024exploiting, zhang2024forget}. Reliability-focused studies examine post-unlearning stability, ensuring that model behavior remains consistent after deletions and that forgotten content does not resurface during downstream use~\cite{schioppa2024model, gao2024meta}. These considerations extend across modalities, including speech and audio systems, where unlearning supports reliable operation after removal of outdated or sensitive data~\cite{cheng2025speech}.

\textbf{Personalization and Preference Control.} Personalization and preference control study how multimodal systems revise or remove user-specific styles, identities, or preferences without retraining core models. In recommendation settings, preference-level unlearning updates user histories or removes modality-specific interactions under legal or licensing constraints while preserving recommendation quality~\cite{sinha2025multi}. VLMs further support lightweight preference control through in-context mechanisms that steer visual behavior at inference time without permanently altering general capabilities~\cite{zhou2024visual}. In text-to-image diffusion, unlearning enables users to suppress unwanted styles or concepts and to prevent reproduction of personalized attributes while maintaining generation fidelity~\cite{biggs2024diffusion, li2024get, polowczyk2025unguide}.

\textbf{Supply-Chain and Backdoor Security.} Supply-chain and backdoor security applications use unlearning to remove malicious associations introduced through poisoned data, hidden triggers, or unsafe fine-tuning, ensuring that released multimodal encoders and generators remain trustworthy in downstream use. In contrastive VLMs, unlearning mitigates poisoning and backdoor threats by weakening or removing learned trigger associations in CLIP-style encoders, improving robustness against malicious training artifacts~\cite{bansal2023cleanclip, liang2024unlearning, liang2024badclip}. 

In diffusion models, unlearning addresses supply-chain risks arising from prompt triggers, spatial patterns, and personalization-based attacks by selectively erasing adversarial concepts or trigger pathways while preserving generation quality~\cite{liu2024metacloak, aravindan2025sealing, jha2025backdoor}. Across modalities, robustness-oriented unlearning aims to prevent the reactivation of malicious behavior after deployment or downstream fine-tuning, supporting safer reuse of pretrained models in open ecosystems~\cite{han2025adaptive, li2025towards}.

\section{Detailed Open Challenges}
\label{app:open_challenges}

\textbf{Theoretical Guarantees.} Despite rapid progress, most multimodal unlearning methods remain heuristic and lack formal guarantees of certified deletion, privacy, or legal compliance. In contrastive and vision-language settings, pair-level removal, single-instance deletion, and secure training procedures approximate forgetting but do not provably eliminate the influence of removed data~\cite{cheng2024multidelete, li2024single, liu2024pre, wang2025muc}.

In diffusion and other generative models, unlearning typically suppresses target concepts without proving erasure, and forgotten content may resurface under downstream fine-tuning or prompt variation~\cite{kim2023towards, park2024direct, zhang2024forget, suriyakumar2024unstable}. Attribution and influence estimation tools provide useful diagnostics but offer only approximate evidence rather than certifiable provenance or deletion guarantees~\cite{dai2023training}. Establishing theoretical foundations and verifiable criteria for multimodal unlearning remains an open challenge.

\vspace{1pt}

\textbf{Cross-Modal Generalization.} Many unlearning studies evaluate on narrow model families, datasets, or modalities, which limits conclusions about general multimodal foundation models. In vision-language encoders and Multimodal Large Language Models (MLLMs), evaluations often center on a small set of architectures or controlled setups, such as CLIP- or LLaVA-only case studies, constraining transfer to broader model ecosystems~\cite{li2024single, dontsov2024clear}. Benchmark analyses further show that unlearning performance is highly sensitive to architectural choices, dataset design, and evaluation tasks~\cite{cheng2024mu, liu2025protecting}.

A similar pattern appears in generative settings, where unlearning is frequently tested on a single diffusion backbone or a limited set of concepts, making it unclear whether findings generalize across architectures, resolutions, or domains~\cite{moon2024holistic, li2025sculpting}. Beyond vision, evaluations in audio, speech, and music typically focus on one model family or dataset, leaving open questions about robustness under multilingual, cross-accent, or cross-genre conditions~\cite{kim2025not, koudounas2025alexa}. Establishing evaluation protocols that span architectures, modalities, and realistic deployment settings remains an open challenge.

\vspace{1pt}

\textbf{Evaluation Reliability.} Evaluation reliability remains a major challenge, as many multimodal unlearning studies rely on proxy-based signals, narrow experimental setups, and unstable metrics, which limits confidence in reported gains across modalities. In VLMs and generative models, success is often assessed using automatic judges, detector outputs, or similarity thresholds on small or synthetic benchmarks, making outcomes highly sensitive to evaluation design rather than underlying model change~\cite{poppi2023removing, xing2024efuf, dai2023training}.

These issues extend to safety, copyright, and privacy settings, where detector-driven or stylized benchmarks can introduce bias and fail to capture whether forgotten concepts are truly removed or merely concealed. As a result, unlearning effectiveness is frequently inferred indirectly, and conclusions may not generalize beyond the specific proxies or model configurations used~\cite{moon2024holistic, zhang2024unlearncanvas}. 

\vspace{1pt}

\textbf{Adversarial Robustness.} Unlearning attempts to erase harmful behavior; however, adversarial robustness remains limited, as backdoors, jailbreaks, and other attack vectors can bring back or bypass forgotten content. In multimodal contrastive learning, existing backdoor and data-protection methods often fail under adaptive threat models, indicating that erased associations may persist in latent representations~\cite{bansal2023cleanclip, zhang2023unlearnable, liu2024multimodal, liang2024badclip}. Diffusion-based text-to-image models exhibit similar fragility: safety-driven unlearning can be bypassed by red-teaming prompts or downstream finetuning, and subject or Not Safe For Work (NSFW) suppression may either miss indirect cues or degrade benign generation when detectors are biased~\cite{kumari2023ablating, park2024direct, liu2024metacloak, chen2025score}.

Black-box and transfer-based attacks further reveal residual traces of supposedly forgotten concepts, suggesting that many unlearning methods attenuate surface behavior rather than fully removing underlying representations~\cite{han2024probing, dang2025diffzoo}. Overall, current defenses trade off safety and utility but remain vulnerable to adaptive reuse, highlighting the need for robustness guarantees that extend beyond static threat assumptions~\cite{huang2024freezeasguard, yoon2025safree, han2025adaptive, li2025towards}.

\vspace{1pt}

\textbf{Utility Trade-offs.} Unlearning often improves safety or compliance at the cost of utility on retained data, neighboring concepts, or benign inputs. In encoder-based models and VLMs, approaches such as CLIP hardening, pair-level deletion, and fine-grained unlearning reduce clean accuracy and cross-dataset transfer, while successful deletion does not guarantee preservation of non-target associations~\cite{bansal2023cleanclip, cheng2024multidelete, li2024single}. Multitask evaluations further indicate that even small deletion ratios can induce measurable performance degradation across modalities~\cite{cheng2024mu}.

In generative models, this trade-off becomes more visible. Stronger forgetting often distorts related styles or reduces visual fidelity, while safety-oriented controls risk over-suppressing benign content or degrading unrelated generations~\cite{kumari2023ablating, liu2024metacloak, han2025adaptive}. These effects reveal a fragile balance between deletion efficacy and utility preservation.

Beyond output quality, unlearning also incurs nontrivial computational cost, which further constrains practical deployment. Many methods require retraining large backbones, maintaining multiple checkpoints, or relying on auxiliary modules and repeated sampling, increasing both compute and storage overhead~\cite{kim2024negmerge, dai2023training, biggs2024diffusion}. Inference-time controls introduce additional latency through extra activations or multiple denoising passes~\cite{cywinski2025saeuron, polowczyk2025unguide}. 

\vspace{1pt}

\textbf{Unified Benchmarks.} Multimodal unlearning still lacks unified benchmarks, as existing evaluations are fragmented, synthetic, or tightly coupled to specific model families. Current suites for VLMs, MLLMs, and speech systems often evaluate a limited set of architectures using synthetic identities, static images, or retrained gold references, making results highly sensitive to model choice, dataset construction, and deletion order~\cite{cheng2024mu, ma2024benchmarking, xu2025pebench, liu2025protecting, koudounas2025alexa}. 

For generative diffusion models, benchmarks typically center on selected concept families or Stable Diffusion-based setups and rely on proxy metrics such as CLIP or Inception scores, which complicates comparison across architectures and limits cross-method reproducibility~\cite{zhang2024unlearncanvas, moon2024holistic, sharma2024unlearning}. 

\section{Disclosure of AI-Assisted Tools}
The authors used Cursor\footnote{\url{https://cursor.com/}} to assist with code development and Grammarly\footnote{\url{https://grammarly.com/}} to support proofreading and language polishing. All inputs were provided by the authors, and all outputs were carefully reviewed and revised.

\end{document}